\pdfoutput=1

\documentclass[11pt]{article}

\usepackage[]{acl}

\usepackage{times}
\usepackage{latexsym}

\usepackage[T1]{fontenc}

\usepackage[utf8]{inputenc}

\usepackage{microtype}
\usepackage{longtable}
\usepackage{ragged2e}
\usepackage{booktabs}
\usepackage{colortbl}
\usepackage{xcolor}
\usepackage{array}
\usepackage{float}
\definecolor{teal}{RGB}{0,128,128}
\definecolor{prowessgray}{gray}{0.93}
\usepackage{inconsolata}
\usepackage{longtable}
\usepackage{ragged2e}
\usepackage{booktabs}
\usepackage{colortbl}
\usepackage{xcolor}
\usepackage{array}
\usepackage{graphicx}
\usepackage{amsmath}
\usepackage{amssymb} 
\usepackage{csquotes}
\usepackage{adjustbox}
\usepackage{xcolor}
\usepackage{booktabs,tabularx}
\usepackage{enumitem} 
\usepackage[T1]{fontenc}

\usepackage{booktabs}
\usepackage{multirow}
\usepackage{makecell}
\usepackage{threeparttable}
\usepackage[table]{xcolor}
\usepackage{tcolorbox}
\usepackage{lipsum} 
\usepackage{setspace}
\definecolor{lightgray}{gray}{0.8} 
\definecolor{pers}{RGB}{230,245,255} 
\definecolor{abn}{RGB}{255,245,230}  

\usepackage{fancyhdr}

\title{\includegraphics[width=0.08\textwidth,trim={50 5 50 5}, clip]{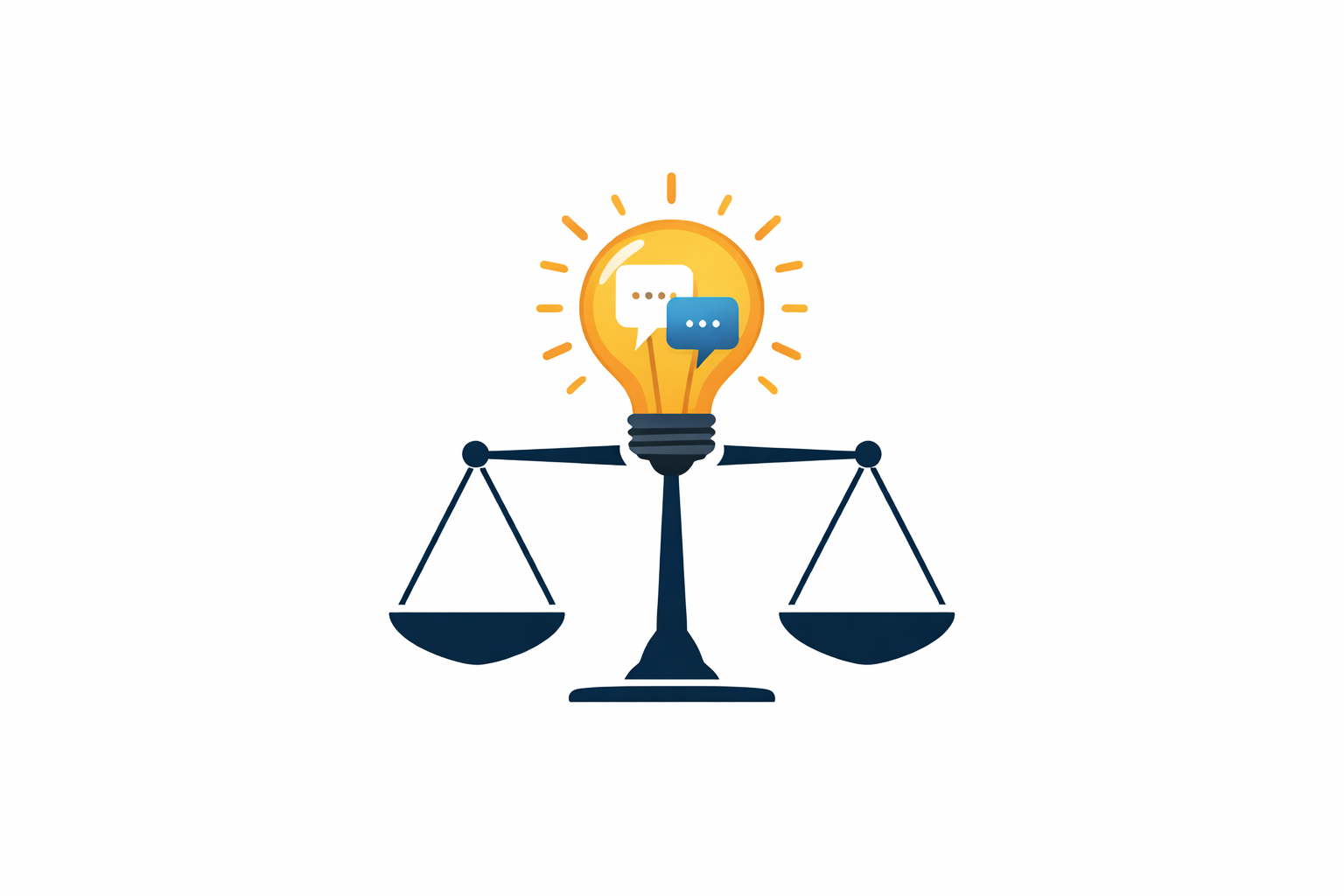} \texttt{DIPLOMAT}: Dialogue-Span-Aware Direct Preference Optimization for Polite Persuasive Workplace Negotiation Dialogues}
\author{Bibhuti Jha$^{\dagger}$\thanks{Authors are jointly first authors.}, Rishikant Chigrupaatii$^{\dagger\text{\textcolor{black}{*}}}$, Priyanshu Priya$^\ddagger$, \textbf{Asif Ekbal}$^\dagger$  \\
$^\dagger$Department of Computer Science and Engineering, Indian Institute of Technology Patna, India  \\
        $^\ddagger$GREYC Laboratory, University of Caen Normandy, France\\
        \texttt{$^\ddagger$\{jhabk369,rishikantchigrupaatii.24,priyanshu528priya}\}{\tt @gmail.com},\\\texttt{$^\dagger$\{bibhuti\_2301cs94,rishikant\_2101cs66,asif\}}{\tt @iitp.ac.in}}

\begin{document}
\maketitle
\begin{abstract}
Effective workplace negotiation requires balancing multiple objectives, including achieving task goals, preserving professional relationships, and resolving conflicts constructively. However, misunderstandings, misaligned preferences, and interpersonal friction often impede successful outcomes. Politeness mitigates these challenges by fostering trust, reducing tension, and preventing escalation, and persuasive communication helps overcome resistance, align preferences, and guide participants toward mutually beneficial agreements. Motivated by these insights, we present \texttt{DIPLOMAT}, a dialogue system for polite and persuasive workplace negotiation. To support its development, we introduce \texttt{PROWESS}, a dataset of multi-turn workplace negotiation dialogues generated via a multi-agent framework and enriched with negotiation strategies, politeness levels, and persuasive strategies. \texttt{DIPLOMAT} is trained using Dialogue-Span-Aware Direct Preference Optimization (DSA-DPO), a novel preference learning objective that identifies key dialogue spans for preference alignment. This enables \texttt{DIPLOMAT} to generate contextually coherent responses that employ intended negotiation strategies, maintain politeness, and incorporate effective persuasion strategies throughout interactions. Automatic and human evaluation on \texttt{PROWESS} confirm that \texttt{DIPLOMAT} consistently outperforms baselines in generating coherent, polite, and persuasive negotiation responses\footnote{Data and code are available at \url{https://github.com/priyanshu-profile/DIPLOMAT} or \url{https://www.iitp.ac.in/~ai-nlp-ml/resources.html\#diplomat}.}. 

\end{abstract}

\section{Introduction}
\begin{figure}[!h]
    \centering
    \includegraphics[width=\linewidth,trim={0 0 0 0},scale = 1.0]{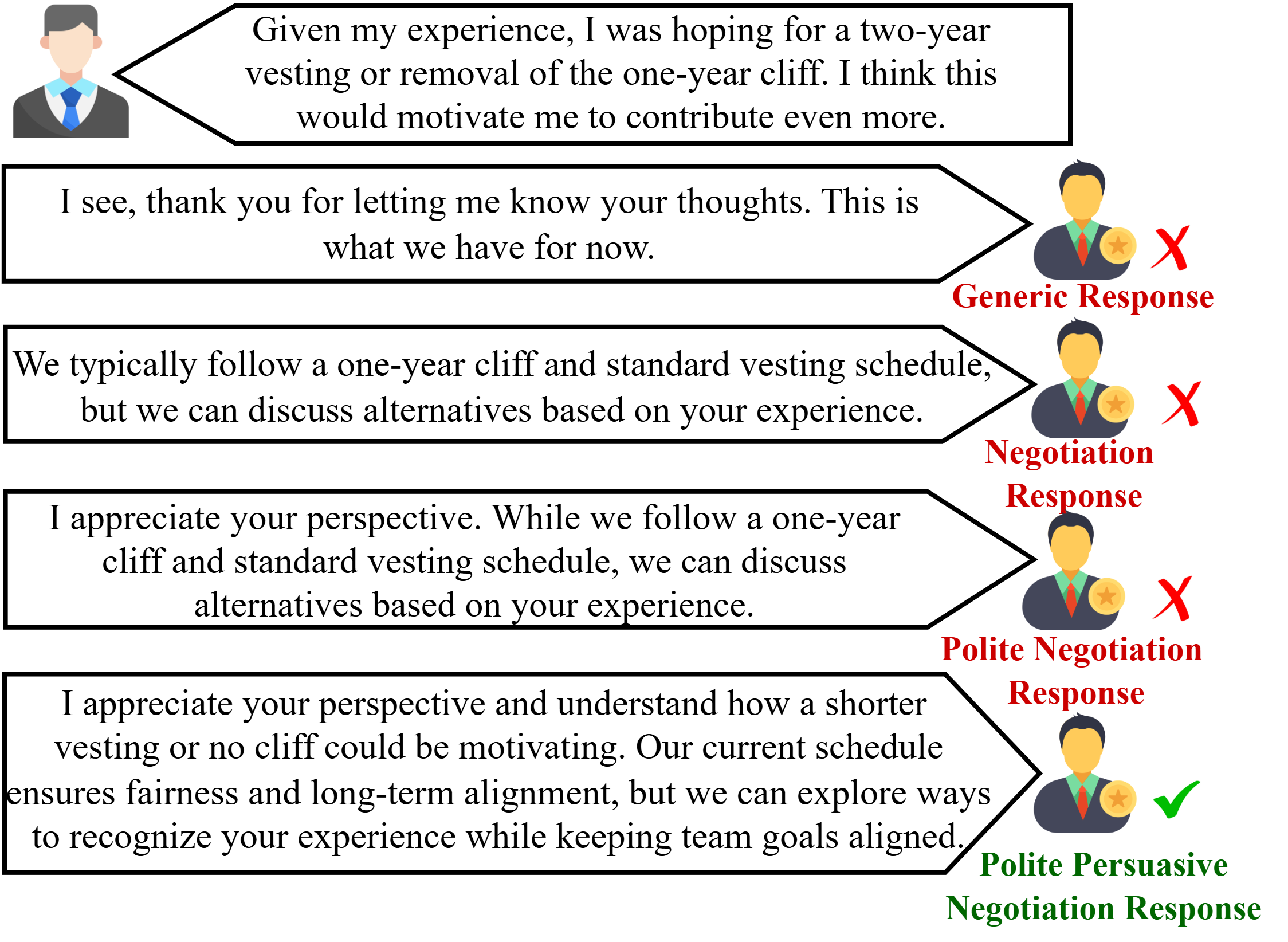}
    \caption{A conversation snippet demonstrating the use of appropriate negotiation strategy (\textit{active listening}), politeness level (\textit{moderate}), and persuasion strategy (\textit{concern addressing}) during workplace negotiation to improve the final outcome.}
    \label{motivation}
\end{figure}

Negotiation dialogue systems are increasingly important in professional settings \cite{zhan-etal-2024-lets}, where agents must not only achieve task goals but also sustain long-term working relationships \cite{lewis-etal-2017-deal, he-etal-2018-decoupling}. Workplace negotiations involve balancing goal achievement, relational harmony, and constructive conflict resolution \cite{carnevale2003negotiation}. Recent advances in large language models (LLMs) \cite{openai2024gpt4technicalreport} and preference-based optimization \cite{rafailov2024directpreferenceoptimizationlanguage} enable fluent, contextually relevant dialogue, yet effectively integrating strategic negotiation behavior with politeness and persuasion remains challenging.

Workplace negotiation is inherently complex due to misaligned preferences, power asymmetries, and interpersonal sensitivities \cite{nash1950bargaining,adair2001negotiation}. Thus, focusing solely on short-term gains can damage relationships, while excessive accommodation may yield poor outcomes. Politeness mitigates friction, fosters trust, and prevents escalation during negotiation \cite{priya2025genteel}, whereas persuasion helps justify proposals, address objections, and gradually align counterparts' decisions \cite{wang-etal-2019-persuasion}. Despite their complementary roles, the interplay between politeness and persuasion remains largely underexplored in existing negotiation dialogue systems. Figure \ref{motivation} illustrates how politeness and persuasion integration during negotiation improves response quality and supports constructive, collaborative outcomes.

Motivated by these considerations, we introduce \texttt{DIPLOMAT}, a polite, persuasive workplace negotiation dialogue system. \texttt{DIPLOMAT} leverages Dialogue-Span-Aware Direct Preference Optimization (DSA-DPO), a novel modification of Direct Preference Optimization (DPO) \cite{rafailov2024directpreferenceoptimizationlanguage} that enables fine-grained preference alignment across multi-turn negotiations. Existing approaches operate either at the utterance or dialogue level. The utterance-level DPO \cite{rafailov2024directpreferenceoptimizationlanguage} optimizes individual responses using positive–negative pairs and can miss evolving negotiation strategies, polite framing, and persuasive justifications across utterances. Dialogue-level DPO \cite{shi-etal-2024-direct,song2024trial} captures long-term patterns but is coarse, potentially obscuring critical contributions, penalizing correct behaviors, and introducing noise, making it harder to identify actions that drive successful negotiation outcomes.

To address these limitations, we propose Dialogue-Span-Aware Direct Preference Optimization (DSA-DPO), a novel approach that captures fine-grained polite and persuasive behaviors across multi-turn negotiations. DSA-DPO identifies key dialogue spans that most strongly influence successful outcomes, rather than optimizing entire dialogues or single utterances. It detects the first critical utterance where negotiation errors occur, samples the preceding interaction history, and forms aligned positive and negative spans of equal length to compute a rigorous preference signal. By focusing on these spans, DSA-DPO reduces noise from non-critical utterances, limits irrelevant variations, and reinforces intended polite and persuasive behaviors that drive successful negotiation outcomes, enabling \texttt{DIPLOMAT} to generate responses that balance task success with relationship maintenance.

To support this research, we introduce \texttt{PROWESS}, a multi-turn dataset of workplace negotiation dialogues generated via a multi-agent framework \cite{fu2023improvinglanguagemodelnegotiation}. In this framework, agents are assigned specialized roles for scenario modeling, dialogue synthesis, and annotation, ensuring logically structured conversations rich in negotiation techniques, polite behaviors, and persuasion tactics that reflect natural negotiation dynamics. Using this framework, we generate 2,400 diverse, high-quality dialogues annotated with negotiation strategies, politeness levels, and persuasion strategies at the utterance level with minimal human supervision.

In summary, the key contributions are: (i) Examine how the dialogue agent’s use of politeness and persuasion influences negotiation outcomes. To our knowledge, this work is the first to model and analyze the combined effects of politeness and persuasion within workplace negotiation context; (ii) Introduce \texttt{DIPLOMAT}, a polite and persuasive negotiation dialogue system that integrates strategic negotiation behavior with politeness and persuasion modeling in workplace interactions; (iii) Propose Dialogue-Span-Aware Direct Preference Optimization (DSA-DPO), a novel preference learning objective that optimizes key dialogue spans, enabling effective alignment of polite and persuasive behaviors with successful negotiation outcomes in \texttt{DIPLOMAT}; (iv) Construct \texttt{PROWESS}, a workplace negotiation dialogue dataset generated via a multi-agent framework and enriched with politeness levels, persuasion and negotiation strategies.

\section{Related Work}
Negotiation dialogue systems have evolved from early agent-based designs for constructing coherent joint plans \cite{mceleney2003agent} and end-to-end neural models for item-division tasks \cite{lewis-etal-2017-deal} to modular strategy generation frameworks \cite{he-etal-2018-decoupling}, persuasion-annotated corpora \cite{chawla-etal-2021-casino}, and reward-based integrative agents \cite{ahmad-etal-2023-ina}. Modeling opponent strategies and personality types improves adaptation \cite{yang-etal-2021-improving}, yet most systems target consumer or e-commerce domains. In workplace negotiation, datasets such as NegoChat \cite{konovalov-etal-2016-negochat} and interview corpora \cite{yamaguchi-etal-2021-dialogue,mannekote2023agreementtrackingmultiissuenegotiation} provide realistic settings, but template-based generation or limited rephrasing still results in dialogues lacking strategic richness. Recent LLM-based efforts explore coaching or multi-agent interactions \cite{shea2024acellmbasednegotiationcoaching,fu2023improvinglanguagemodelnegotiation,abdelnabi2024cooperationcompetitionmaliciousnessllmstakeholders}, yet primarily optimize deal-making metrics rather than socially-aware negotiation behaviors.

Politeness \cite{brown1987politeness} is a key social driver in negotiation that fosters rapport, trust, and cooperation \cite{lee2021beyond,monson2022pampering}, and has been shown to improve negotiation outcomes across diverse settings \cite{10.1145/3472306.3478336,10.1371/journal.pone.0212306,priya2025genteel}. Persuasion \cite{eagly1984cognitive} similarly plays a central role in negotiation by shaping how proposals are framed, justified, and defended to influence counterpart decisions \cite{ledgerwood2006changing,provis2004negotiation}. Persuasion enables the participants to advocate for their interests while guiding the dialogue toward mutually acceptable terms \cite{wang-etal-2019-persuasion,dutt-etal-2021-resper}. 

Recently, preference learning has become a dominant paradigm for aligning LLMs with desired behaviors. Methods progress from Reinforcement Learning from Human Feedback (RLHF) \cite{ouyang2022traininglanguagemodelsfollow} to Direct Preference Optimization (DPO) \cite{rafailov2024directpreferenceoptimizationlanguage} and its dialogue-aware variants \cite{shi-etal-2024-direct,song2024trial,jung-etal-2025-diatool,zhuang2025teaching,hama-etal-2025-rapsil}. Despite extensive research on negotiation dialogue agents, existing systems primarily focus on modeling deal-making strategies or opponent behaviors and are largely developed for consumer-facing domains. While prior studies highlight the importance of social signals such as politeness and persuasion for successful negotiation, these aspects are rarely modeled jointly in automated negotiation dialogue systems. Furthermore, current alignment methods for dialogue agents rely on preference learning at the utterance or dialogue level, which often fails to capture nuanced negotiation dynamics across multi-turn interactions. To address these limitations, we introduce \texttt{DIPLOMAT}, a dialogue-span-aware DPO framework that leverages fine-grained preference signals over politeness, persuasion, and negotiation dynamics to align LLMs for generating polite and persuasive workplace negotiation dialogues. \texttt{DIPLOMAT} is designed to be scalable and adaptable across diverse application domains. 

\section{Dataset}
To evaluate \texttt{DIPLOMAT}, we introduce \texttt{PROWESS}, a novel dialogue dataset designed for workplace negotiation scenarios. \texttt{PROWESS} consists of multi-turn interactions between a candidate and an employer, covering key negotiation dimensions such as compensation, working conditions, career progression, and related employment terms. The dataset consists of 2,400 dialogues spanning 25 common workplace negotiation aspects, capturing realistic negotiation behaviors and conversational dynamics commonly observed during hiring and professional interactions in the workplace. The list of negotiation aspects in the \texttt{PROWESS} dataset is provided in Table~\ref{tab:prowess_aspects}. 

\begin{table}[H]
\centering
\small
\begin{tabular}{p{7.5cm}}
\toprule
\textbf{Negotiation Aspects} \\
\midrule
benefits negotiation, bonus structure, career development, contract terms, equity negotiation, exit interview, flexible hours, health benefits, hiring decision, job responsibilities, onboarding, performance review, professional development, project assignment, promotion discussion, relocation package, remote work arrangement, retention offer, retirement plans, salary negotiation, team transition, training opportunities, vacation time, work equipment, work-life balance \\
\bottomrule
\end{tabular}
\caption{Negotiation aspects covered in the \texttt{PROWESS} dataset.}
\label{tab:prowess_aspects}
\end{table}

\subsection{Dataset Construction}
\begin{figure}[!h]
    \centering
    \includegraphics[width=\linewidth,trim={0 0 0 0},scale = 1.0]{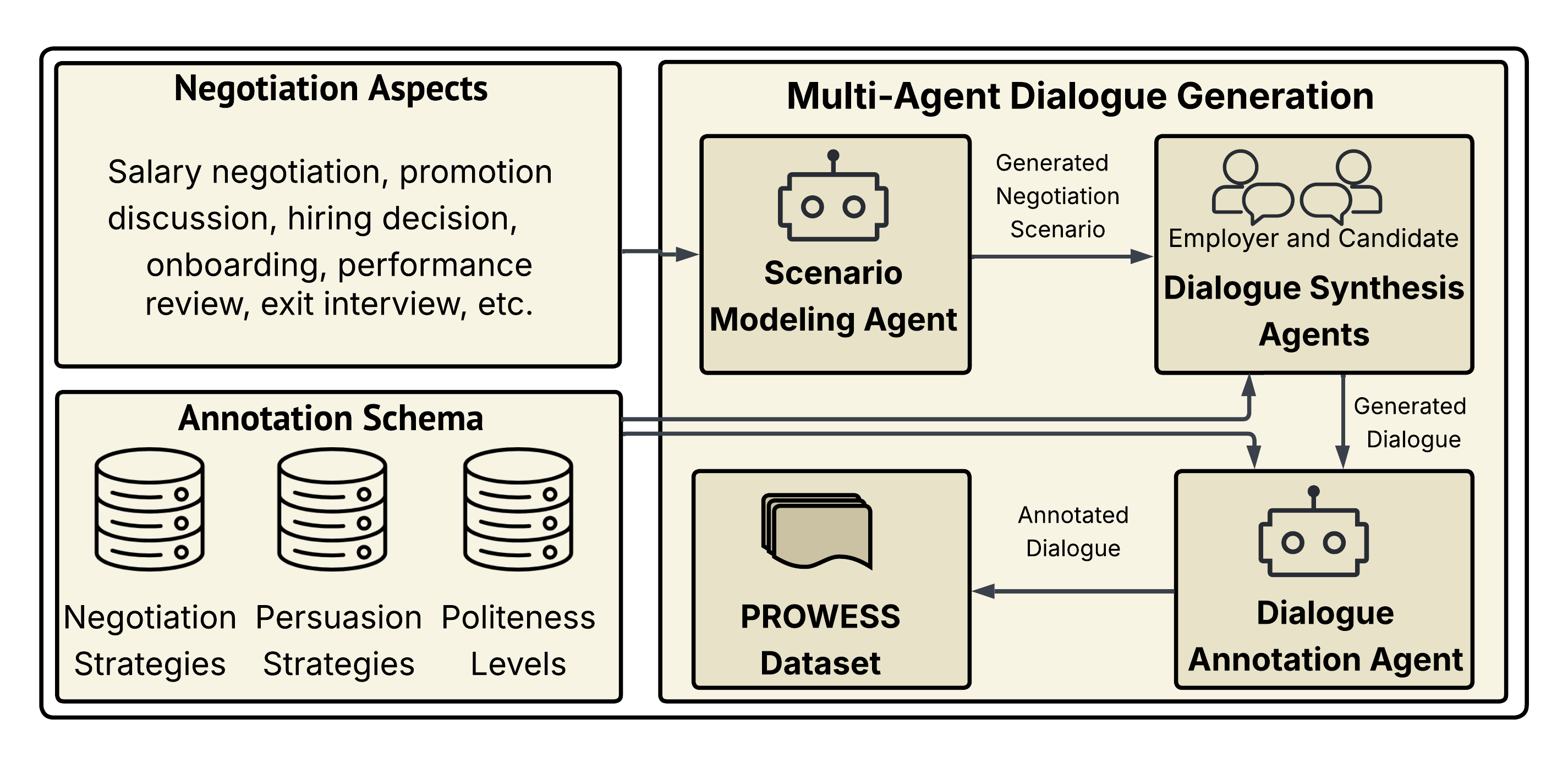}
    \caption{\texttt{PROWESS} dataset construction pipeline.}
    \label{data_pipeline}
\end{figure}

The dataset construction involves two steps: (1) Negotiation, persuasion, and politeness annotation schema designing and (2) Multi-agent dialogue generation. Figure~\ref{data_pipeline} illustrates the \texttt{PROWESS} dataset construction pipeline. 

\paragraph{(1) Negotiation, Persuasion, and Politeness Annotation Schema Designing.}
Workplace negotiation relies on information sharing, transparency, and relational rapport to support collaboration and professional growth. Effective negotiation strategies are essential for achieving mutually beneficial outcomes, resolving conflicts, and maintaining professional relationships. Considering this, we define a set of 11 negotiation strategies grounded in workplace negotiation principles \cite{thompson2005mind}. These strategies are designed for problem-solving, concession-making, position-establishing, or relationship-building, and include \textit{collaborative style, active listening, win-win framing, principled negotiation, data-driven justification, MESO (Multiple Equivalent Simultaneous Offers), anchoring, door-in-the-face, reciprocal concessions, credibility assertion,} and \textit{no strategy}.

While negotiation strategies provide the foundation for effective bargaining, politeness and persuasion complement them by guiding negotiation dynamics, influencing decisions, and shaping relationships. Politeness, in particular, influences negotiation outcomes by mitigating conflict and promoting sociopsychological closeness \cite{brown1987politeness}. Users express varying levels of politeness, which shape how negotiations unfold over time \cite{OVERBECK2010126}. In practice, we adopt three politeness level categories from workplace negotiation practice \cite{yuxian2013politeness}, including \textit{low\_polite, moderate\_polite,} and \textit{high\_polite}. Persuasion further enables alignment of preferences and facilitates agreement in situations with asymmetric power, conflicting goals, or varying engagement. Guided by the persuasive negotiation theory \cite{fowler1998negotiating}, we define a set of 11 persuasion strategies: 
\textit{rapport building, concern addressing, emotional appeal, credibility \& confidence, data-driven persuasion, problem-solving focus, self-interest appeal, value alignment, reputation highlighting, future vision alignment}, and \textit{no strategy}. Definitions and examples of negotiation strategies, politeness levels, and persuasion strategies are given in Appendix \ref{dataset_details}.

\paragraph{(2) Multi-Agent Dialogue Generation.}
We construct the \texttt{PROWESS} dataset using a structured multi-agent framework that coordinates specialized LLM agents to generate negotiation dialogues. The framework consists of three agents, \textit{viz.} scenario modeling, dialogue synthesis, and dialogue annotation, each responsible for a specific task. Each agent operates under shared scenario constraints and role-specific objectives, ensuring that the output of one agent reliably conditions the next agent's behavior, resulting in coherent, goal-driven negotiation dialogues. This integrated modular design allows precise control over dialogue quality and annotation consistency. The scenario modeling agent employs Gemini-2.5-Pro \cite{comanici2025gemini}, and dialogue synthesis and dialogue annotation agents employ GPT-4o-mini \cite{openai2024gpt4technicalreport} for its strong reasoning and contextual tracking capabilities, enabling reliable guidance throughout the data generation process. 

\textbf{Scenario Modeling Agent.} The Scenario Modeling Agent creates structured negotiation setups by modeling key elements, namely negotiation aspect, background context, employer and candidate roles, negotiation goal, and candidate starting position. This ensures scenarios are realistic and diverse, covering varied objectives, roles, and initial conditions. For each negotiation aspect, 16 distinct scenarios are generated to support the generation of large-scale, high-quality negotiation dialogues.

\textbf{Dialogue Synthesis Agents.} The Dialogue Synthesis Agents generate multi-turn negotiation dialogues by coordinating two LLM agents representing the employer and the candidate, whose behavior is grounded in the role-specific objectives and constraints defined in the structured scenario generated by the Scenario Modeling Agent. Building on prior work on multi-agent dialogue simulation \cite{fu2023improvinglanguagemodelnegotiation}, the agents generate conversations by cyclically feeding the output of one agent as input to the other to generate coherent, multi-turn interactions. To ensure diverse and realistic conversational behavior, the interaction is guided by structured prompts specifying negotiation and persuasion strategies for each agent. The agents are further instructed to dynamically adapt their strategies in response to the other participant’s behavior. Additionally, both agents are briefed to maintain a professional and polite tone aligned with workplace norms, while adjusting their tone to reflect the interaction dynamics. The dialogues commence when the employer agent is prompted with \enquote{\textit{Start the conversation}}, initiating a structured, multi-turn negotiation exchange that reveals each participant’s adaptive negotiation and persuasion strategies, as well as professional, polite behavior. 

\textbf{Dialogue Annotation Agent.} After each round of conversation (a pair of turns between the employer and candidate agents), the Dialogue Annotation Agent first evaluates participants’ utterances to annotate politeness levels. It then annotates the persuasion and negotiation strategies used, along with a brief rationale for each strategy annotation. To ensure consistency across rounds, both politeness and strategy annotations are drawn from predefined categories. The agent considers the cumulative dialogue context to capture the evolution of strategies and polite tone. To guide its behavior and minimize annotation errors, we provide the agent with three examples for each annotation dimension. These annotations facilitate analysis of negotiation strategy adaptation, persuasive effectiveness, and politeness management, and provide supervision for training models that generate polite and persuasive negotiation dialogues. 

The prompt templates for scenario modeling, dialogue synthesis, and dialogue annotation agents are detailed in Appendix~\ref{subsec:prompt-templates}. 

\subsection{Dataset Filtering and Quality Assessment}
Once the complete dataset with 2,675 dialogues is generated, we perform a comprehensive inspection to remove erroneous dialogues. Specifically, we identify five types of issues: (i) empty utterances, (ii) repetitive utterances, (iii) insufficient interaction rounds, (iv) incomplete or missing annotations, and (v) improper conversation openings or closings. The dialogues exhibiting any of these issues are removed to improve overall dataset quality and reliability. 

The remaining dialogues are qualitatively evaluated by the three human evaluators\footnote{The evaluators hold Ph.D. degrees in Linguistics and have expertise in negotiation, persuasion, and politeness concepts. They are compensated in as per the institute guidelines.} under the supervision of a domain expert in business management to ensure contextual accuracy and domain relevance. Each dialogue is rated for Scenario Consistency (SC), Negotiation Strategy Correctness (NSC), Persuasion Strategy Correctness (PSC), Politeness Appropriateness (PA), Fairness (F), Coherence (C), Naturalness (N), and Engagingness (E) (Descriptions are given in Appendix \ref{dataset_details}), using a 1–5 scale (low to high). We retain only those dialogues that receive scores of $\geq 3$ across all evaluation metrics. The retained dialogues achieve average ratings of 4.58 (SC), 4.46 (NSC), 4.39 (PSC), 4.51 (PA), 4.32 (F), 4.37 (C), 4.79 (N), and 4.41 (E), with corresponding inter-evaluator $\kappa$ scores of 0.84, 0.82, 0.80, 0.83, 0.79, 0.81, 0.85, and 0.80 for SC, NSC, PSC, PA, F, C, N, and E, respectively. The $\kappa$ values indicate substantial to near-perfect inter-evaluator agreement, confirming the reliability and consistency of the human evaluations. 

Overall, these results demonstrate that the dialogues exhibit strong scenario grounding, consistent use of negotiation and persuasion strategies, appropriate politeness, and high levels of fairness, coherence, naturalness, and engagement. The final dataset statistics are reported in Table \ref{tab:dataset_stats}.

\begin{table}[ht]
\centering
\begin{tabular}{l|r}
\hline
\textbf{Statistic} & \textbf{Value} \\
\hline
\# Negotiation Aspects & 25 \\
\# Dialogues & 2,400 \\
\# Utterances & 51,796 \\
\# Words & 941,900 \\
Avg. Utterances per Dialogue & 21.58 \\
Minimum Dialogue Length & 18 \\
Maximum Dialogue Length & 26 \\
Avg. Words per Dialogue & 392.5 \\
Avg. Words per Utterance & 18.2 \\
\hline
\end{tabular}
\caption{\texttt{PROWESS} dataset statistics.}
\label{tab:dataset_stats}
\end{table}

\section{Methodology}
\begin{figure*}[!h]
    \centering
    \begin{adjustbox}{max width = 0.9\linewidth}
    \includegraphics[trim={0 0 0 0},scale = 1.0]{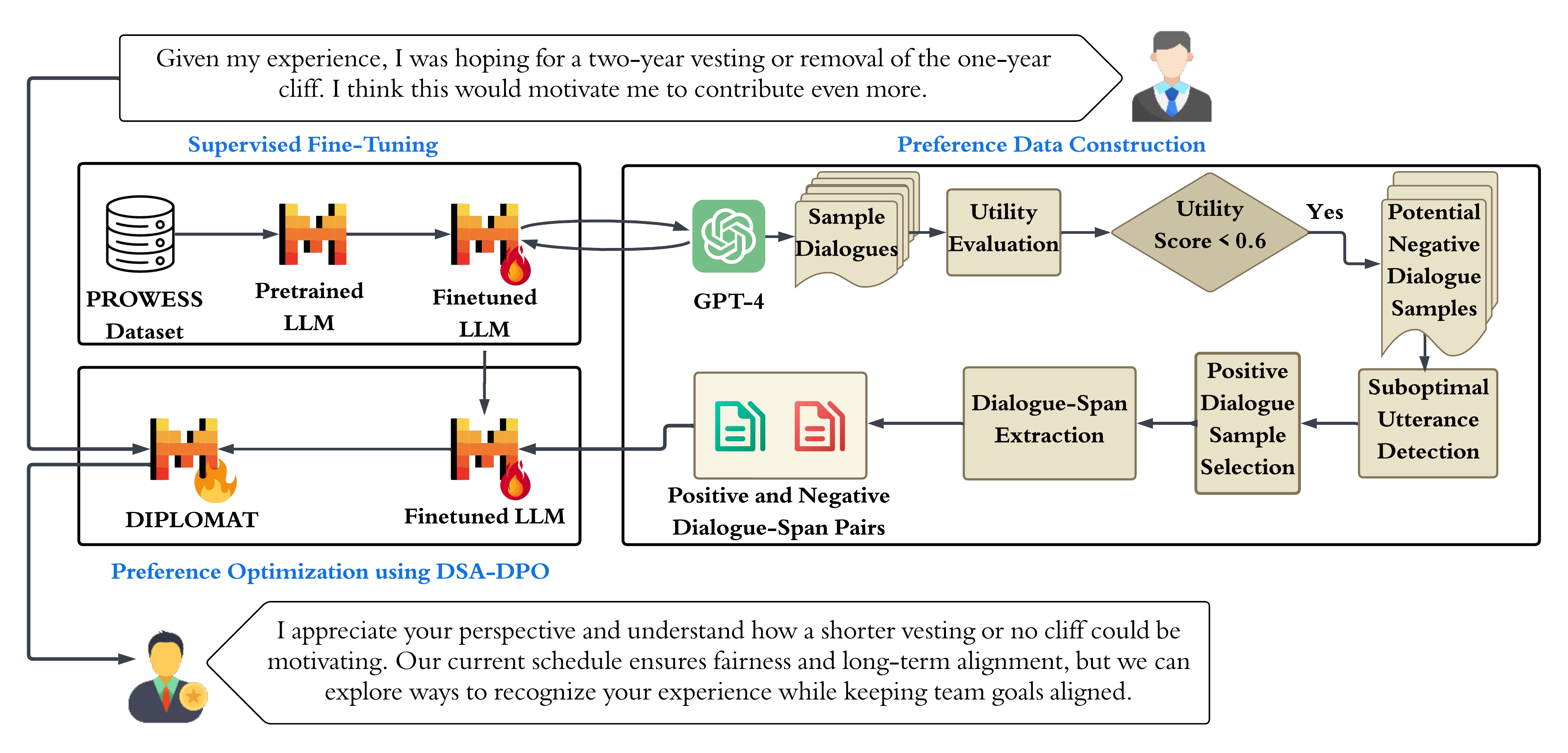}
    \end{adjustbox}
    \caption{Architecture of the proposed polite persuasive negotiation dialogue agent, \texttt{DIPLOMAT}. \texttt{DIPLOMAT} is developed in three stages: (1) \textit{Supervised Fine-Tuning (SFT)}, where the pre-trained LLM is fine-tuned on negotiation dialogues from the \texttt{PROWESS} dataset to obtain the supervised fine-tuned LLM ($\pi_\text{SFT}$), which generates polite and persuasive negotiation responses; (2) \textit{Preference Data Construction}, where $\pi_\text{SFT}$  interacts with GPT-4o-mini to sample diverse negotiation dialogues, which are evaluated using a utility function. Dialogues with utility score $< 0.6$ are treated as potential negative samples, which are then paired with positive dialogue samples through suboptimal utterance detection, positive dialogue sample selection, and dialogue-span extraction, yielding preference pairs of positive and negative dialogue spans; and (3) \textit{Preference Optimization using DSA-DPO}, where $\pi_\text{SFT}$ is further optimized using the proposed DSA-DPO objective over the extracted positive and negative dialogue spans to generate contextually coherent, polite, persuasive, and strategically consistent negotiation responses.}
    \label{diplomat_arch}
\end{figure*}

The proposed \texttt{DIPLOMAT} framework consists of three stages: (1) supervised fine-tuning on the \texttt{PROWESS} dataset, (2) preference data construction through utility evaluation, suboptimal utterance detection, positive dialogue sample selection, and dialogue-span extraction, and (3) preference optimization using dialogue span-aware DPO (DSA-DPO) objective. Figure~\ref{diplomat_arch} provides an overview of \texttt{DIPLOMAT}.

\subsection{Supervised Fine-Tuning}
Let $\mathcal{D} = \{d_j\}_{j=1}^{N}$ denote a dataset of $N$ dialogues. For the $t$-th turn of the $j$-th dialogue, the conversational context $\mathcal{C}_t = (e_1, c_1, \dots, e_{t-1}, c_{t-1}, c_t) \in d_j$ comprises alternating employer ($e$) and candidate ($c$) utterances upto $(t-1)$ turns, together with the candidate’s current utterance $c_t$. The primary objective is to generate the employer's appropriate response $e_t$ conditioned on $\mathcal{C}_t$. 

To generate contextually appropriate responses, the employer must reason about the candidate’s polite, persuasion, and negotiation behaviors. Concretely, given a context $\mathcal{C}_t$, the model is expected to: (i) infer candidate’s politeness level, and persuasion and negotiation strategies along with their rationale, (ii) select appropriate negotiation and persuasion strategies for employer based on candidate's strategies and adjust employer’s politeness level accordingly; and (iii) generate employer's final response conditioned on these inferred strategies and politeness level. These intermediate reasoning steps provide additional supervision that facilitates learning of strategy adaptation, politeness mirroring, conflict de-escalation, and effective concession timing during negotiation. To operationalize this, we construct structured training target sequences with three components: (a) \textit{candidate inference block} describing candidate negotiation and persuasion strategies with reasoning, and the candidate’s politeness level, (b) \textit{employer inference block} specifying employer’s selected negotiation and persuasion strategies with reasoning, and employer’s politeness level, and (c) \textit{employer response block} outlining employer's final response.

During supervised fine-tuning, the model is trained to generate these structured components sequentially before generating the final response. Specifically, we fine-tune the pre-trained LLM \texttt{Mistral-7B-Instruct} \cite{jiang2023mistral7b} on $\mathcal{D}$ using standard next-token prediction. Formally, the model parameters $\theta$ are optimized by minimizing the negative log-likelihood loss, yielding supervised fine-tuned model $\pi_{\text{SFT}}$:
\begin{equation}
\label{lsft}
\mathcal{L}_{\text{SFT}}(\theta)
= -\mathbb{E}_{{(\mathcal{C}_t,y_t)\sim \mathcal{D}}}
\left[
\sum_{{i=1}}^{{|y_t|}}
\log \pi_{\theta}(y_{t_i}\mid y_{t_{<i}}, \mathcal{C}_t)
\right]
\end{equation}
where $y_t$ denotes the structured target sequence comprising the candidate inference, employer inference, and employer response blocks. At inference time, these intermediate blocks can be treated as latent and optionally suppressed, allowing the model to generate responses conditioned solely on the dialogue context. This supervised fine-tuning step equips the LLM to understand the candidate’s behavior, adapt its own, and generate contextually appropriate, polite, and persuasive responses during negotiation, serving as the initialization for subsequent preference-based optimization.

\subsection{Preference Data Construction}
The construction of a high-quality dialogue-span-aware preference dataset is central to our approach. Unlike standard preference datasets that treat dialogues as monolithic units, ours captures fine-grained dialogue spans that reflect strategic conversational behaviors, including politeness adaptation, persuasive maneuvers, and negotiation strategy shifts, allowing agents to learn how individual utterances influence overall negotiation outcomes. This span-aware supervision enables the negotiation agent to generate contextually appropriate, polite, and persuasive responses in multi-turn negotiations. To construct the dataset, we first generate multiple negotiation dialogues under a given scenario and goal. Specifically, the supervised fine-tuned model $\pi_{\text{SFT}}$ interacts with GPT-4o-mini to simulate negotiation conversations and generate diverse dialogue samples. These generated dialogues serve as the input to the preference data construction pipeline, where utility evaluation, suboptimal utterance detection, positive dialogue selection, and dialogue-span extraction are performed to construct fine-grained dialogue-span-aware preference pairs.

\textbf{(i) Utility Evaluation.} The generated dialogues are evaluated for quality using a utility score $U(\cdot)$ that captures multiple dimensions of negotiation success: $U(d) =
\text{PT}(d)
+ \text{PSA}(d)
+ \text{NSA}(d) 
+ \text{MS}(d)
+ \text{AQ}(d)
- \text{NB}(d)$.
Here, $\text{PT}(\cdot)$, $\text{PSA}(\cdot)$, $\text{NSA}(\cdot)$, $\text{MS}(\cdot)$, $\text{AQ}(\cdot)$, and $\text{NB}(\cdot)$ denote the scores for \textit{Politeness Trajectory}, \textit{Persuasion Strategy Alignment}, \textit{Negotiation Strategy Alignment}, \textit{Mutual Satisfaction}, \textit{Agreement Quality}, and \textit{Negotiation Breakdown}, respectively, for a dialogue $d$. These scores are estimated by Gemini-2.5-Pro using structured evaluation prompts (Appendix \ref{subsec:prompt-templates}). 

The dialogues with $U(\cdot) < 0.6$ are treated as potential negative samples, indicating suboptimal negotiation outcomes or undesirable conversational behaviors. Each utility component is evaluated in the range $[0,1]$, consequently, $U(\cdot) \in $ $[-1$ to $5]$. Although the overall utility score spans a wider range, the threshold analysis is focused on the interval $[0,1]$ because this region empirically corresponds to suboptimal yet non-trivial negotiation dialogues that provide informative negative samples for preference optimization. Extremely low utility values are rare and typically correspond to severe negotiation breakdowns, while substantially higher values generally indicate dialogues with acceptable negotiation quality that are less suitable as negative samples. A sensitivity analysis on the utility threshold is given in Appendix~\ref{appendix_threshold_analysis}. For each identified negative sample, we then construct dialogue-span-aware preference pairs as follows:

\textbf{(ii) Suboptimal Utterance Detection.} Unlike tasks with objectively defined errors (e.g., math or programming), suboptimal utterances in negotiation dialogues are inherently ambiguous. In a negative dialogue sample, an utterance is identified as suboptimal using Gemini-2.5-Pro via structured prompts if it satisfies one or more utility-based criteria: (a) it is pivotal for advancing the negotiation goal, such as offers, counteroffers, or concessions; (b) it exhibits low effectiveness in one or more utility dimensions: PT, PSA, NSA, MS, or AQ; or (c) it contributes to a negotiation breakdown or reduces relationship quality, indicating a missed opportunity for polite or persuasive engagement. 

\textbf{(iii) Positive Dialogue Sample Selection.} After suboptimal utterance detection, we identify candidate positive dialogues that serve as corrective exemplars for the flagged utterance. To this end, we retrieve three complete dialogues sharing the same conversational context preceding the identified suboptimal utterance. Among these candidates, the dialogue with the highest utility score is selected as the candidate sample. If its utility score exceeds that of the negative dialogue sample, the candidate and negative samples form a preference pair; otherwise, the negative sample is discarded. 

\textbf{(iv) Dialogue-span Extraction.} Once positive and negative dialogue samples are paired, we extract fine-grained dialogue spans that capture the utterances most responsible for the observed differences in utility score. To this end, both dialogues are provided to Gemini-2.5-Pro with structured prompts to identify a contiguous span in the positive sample that maximally contributes to its superior utility across the dimensions of PT, PSA, NSA, MS, AQ, and minimal NB. A span of equal length is then extracted from the corresponding negative dialogue sample, forming a dialogue-span-aware preference pair. This stage filters out irrelevant utterances, such as greetings or pleasantries, ensuring that the model focuses on learning strategically and socially meaningful improvements at the utterance level in multi-turn negotiations. 

The resulting preference dataset comprises 700 pairs. Appendix~\ref{app_pref_data} details the evaluation of Gemini-2.5-Pro within the preference data construction pipeline, and Appendix~\ref{subsec:prompt-templates_pref} provides the prompt templates used in this pipeline.

\subsection{Preference Optimization using DSA-DPO} We leverage a curated dialogue-span-aware preference dataset $\mathcal{D}_{\text{pref}} = \{(\mathcal{C}_s^{m},\tau_w^{m}, \tau_l^{m})\}_{m=1}^{M}$, where $\mathcal{C}_{s}$ denotes the conversational context up to the suboptimal utterance $s$, and $\tau_w$ and $\tau_l$ represent preferred and non-preferred dialogue spans extracted from multi-turn negotiation interactions, to further refine the supervised fine-tuned policy $\pi_{\text{SFT}}$, and $M$ denote the total no. of instances in $\mathcal{D}_{\text{pref}}$. Specifically, we adopt a dialogue-span-aware direct preference optimization (DSA-DPO) objective, which operates on dialogue spans rather than isolated responses. This objective encourages the model to assign higher likelihood to spans exhibiting polite, persuasive, and strategically coherent negotiation behaviors, such as maintaining politeness, providing constructive persuasive arguments, and progressively guiding the conversation toward mutually beneficial agreement, while penalizing spans with undesirable behaviors such as blunt demands or uncooperative tone. Formally, the DSA-DPO objective is defined as:
\begin{equation}
\begin{aligned}
\mathcal{L}_{\text{DSA-DPO}} &= 
- \mathbb{E}_{(\mathcal{C}_s,\tau_w,\tau_l)\sim \mathcal{D}_{\text{pref}}} 
\Bigg[ \log \\
&\quad 
\sigma \Bigg(
\beta \Big(
\log \frac{\pi_\theta(\tau_w)}{\pi_{\text{ref}}(\tau_w)}
-
\log \frac{\pi_\theta(\tau_l)}{\pi_{\text{ref}}(\tau_l)}
\Big)
\Bigg)
\Bigg]
\end{aligned}
\end{equation}
where,
\begin{equation}
\pi_{\theta}(\tau) = \prod_{t=s}^{s+p} \pi_{\theta}(y_t \mid \mathcal{C}_t),
\end{equation}
and $\pi_{\text{ref}}$ denotes the reference policy initialized from $\pi_{\text{SFT}}$, $\sigma$ is the logistic function, $\beta$ controls the strength of preference optimization, and $p$ is the number of utterances in the chosen dialogue spans. By optimizing over multi-turn dialogue spans, DSA-DPO learns how polite and persuasive behaviors evolve across utterances, enabling the model to capture negotiation dynamics and lead to more effective negotiation strategies. 

\begin{table*}[t]
\centering
\small
\begin{adjustbox}{max width=0.9\linewidth}
\begin{tabular}{lccccccccc}
\toprule
\textbf{Models} & \textbf{PPL $\downarrow$} & \textbf{B-4 $\uparrow$} & \textbf{BS-F1 $\uparrow$} & \textbf{D-2 $\uparrow$} & \textbf{R-LEN $\uparrow$} & \textbf{NSC $\uparrow$} & \textbf{PSC $\uparrow$} & \textbf{PA $\uparrow$} & \textbf{AR (\%) $\uparrow$} \\
\midrule
INA & 14.6 & 0.344 & 0.602 & 0.520 & 16.5 & 0.520 & 0.456 & 0.858 & 64.3 \\
\textsc{genteel-negotiator} & 8.7 & 0.428 & 0.598 & 0.615 & 17.0 & 0.592 & 0.536 & 0.875 & 75.7 \\
ProCoT (GPT-5-mini) & 10.4 & 0.351 & 0.575 & 0.575 & 15.4 & 0.562 & 0.492 & 0.865 & 69.4 \\
\midrule
Mistral-7B-Instruct-v0.3 & 22.4 & 0.229 & 0.569 & 0.352 & 14.2 & 0.412 & 0.372 & 0.841 & 60.2 \\
Llama3.1-8B & 31.2 & 0.392 & 0.568 & 0.390 & 13.9 & 0.418 & 0.386 & 0.835 & 34.8 \\
Gemma-2-9B & 11.4 & 0.286 & 0.571 & 0.370 & 15.1 & 0.414 & 0.382 & 0.862 & 32.5 \\
\midrule
Mistral-7B-Instruct-v0.3-SFT & 9.1 & 0.373 & 0.601 & 0.594 & 16.3 & 0.558 & 0.490 & 0.863 & 68.6 \\
Llama3.1-8B-SFT & 10.8 & 0.126 & 0.609 & 0.593 & 17.3 & 0.498 & 0.448 & 0.864 & 66.9 \\
Gemma-2-9B-SFT & 19.8 & 0.368 & 0.607 & 0.613 & 16.4 & 0.498 & 0.446 & 0.865 & 64.1 \\
\midrule
Mistral-7B-Instruct-v0.3-SFT-DPO & 8.8 & 0.407 & 0.606 & 0.615 & 18.5 & 0.598 & 0.528 & 0.869 & 73.4 \\
Mistral-7B-Instruct-v0.3-SFT-DMPO & 8.7 & 0.413 & 0.608 & 0.622 & 17.1 & 0.618 & 0.548 & 0.871 & 76.2 \\
Mistral-7B-Instruct-v0.3-SFT-ETO & 8.7 & 0.434 & 0.612 & 0.625 & 17.2 & 0.624 & 0.552 & 0.872 & 77.5 \\
Mistral-7B-Instruct-v0.3-PositiveSFT & 8.9 & 0.396 & 0.594 & 0.606 & 16.8 & 0.574 & 0.506 & 0.866 & 70.8 \\
\midrule
\rowcolor{lightgray}
\textbf{DIPLOMAT} & \textbf{8.6} & \textbf{0.452} & \textbf{0.630} & \textbf{0.635} & \textbf{19.8} & \textbf{0.724} & \textbf{0.628} & \textbf{0.882} & \textbf{82.3} \\
\bottomrule
\end{tabular}
\end{adjustbox}
\caption{Automatic evaluation results. Results are
statistically significant at 5\% significance level based on t-test \cite{welch1947generalization}.}
\label{tab:automatic_results}
\end{table*}

\begin{table*}[t]
\centering
\small
\begin{adjustbox}{max width=0.85\linewidth}
\begin{tabular}{lcccccccccc}
\toprule
\textbf{Models} & \textbf{F $\uparrow$} & \textbf{CC $\uparrow$} & \textbf{E $\uparrow$} & \textbf{NSC $\uparrow$} & \textbf{PSC $\uparrow$} & \textbf{PA $\uparrow$} & \textbf{BE $\uparrow$} & \textbf{OF $\uparrow$} & \textbf{SC $\uparrow$} & \textbf{IE $\uparrow$} \\
\midrule
INA & 3.27 & 3.11 & 3.18 & 2.58 & 2.46 & 4.25 & 2.88 & 2.69 & 3.01 & 3.14 \\
\textsc{genteel-negotiator} & 4.09 & 4.21 & 4.16 & 3.83 & 3.69 & 4.37 & 3.95 & 3.79 & 3.88 & 4.03 \\
ProCoT (GPT-5-mini) & 3.91 & 3.85 & 3.77 & 3.08 & 2.96 & 4.30 & 3.48 & 3.39 & 3.44 & 3.58 \\ \midrule
Mistral-7B-Instruct-v0.3 & 2.89 & 2.95 & 2.81 & 2.02 & 2.11 & 4.19 & 2.59 & 2.43 & 2.55 & 2.70 \\
Llama3.1-8B & 2.78 & 2.83 & 2.74 & 2.08 & 2.03 & 4.15 & 2.54 & 2.39 & 2.48 & 2.53 \\
Gemma-2-9B & 3.06 & 3.01 & 2.98 & 2.13 & 2.21 & 4.31 & 2.60 & 2.53 & 2.56 & 2.59 \\ \midrule
Mistral-7B-Instruct-v0.3-SFT & 3.79 & 3.89 & 3.72 & 3.02 & 2.97 & 4.31 & 3.49 & 3.38 & 3.43 & 3.55 \\
Llama3.1-8B-SFT & 3.74 & 3.68 & 3.63 & 2.89 & 2.81 & 4.29 & 3.38 & 3.28 & 3.35 & 3.47 \\
Gemma-2-9B-SFT & 3.69 & 3.62 & 3.58 & 2.95 & 2.89 & 4.32 & 3.33 & 3.24 & 3.30 & 3.39 \\ \midrule
Mistral-7B-Instruct-v0.3-SFT-DPO & 4.01 & 4.04 & 3.95 & 3.29 & 3.24 & 4.34 & 3.73 & 3.64 & 3.69 & 3.84 \\
Mistral-7B-Instruct-v0.3-SFT-DMPO & 4.08 & 4.14 & 4.02 & 3.41 & 3.35 & 4.35 & 3.79 & 3.72 & 3.77 & 3.89 \\
Mistral-7B-Instruct-v0.3-SFT-ETO & 4.13 & 4.18 & 4.09 & 3.47 & 3.43 & 4.36 & 3.84 & 3.77 & 3.82 & 3.94 \\
Mistral-7B-Instruct-v0.3-PositiveSFT & 3.88 & 3.94 & 3.83 & 3.14 & 3.08 & 4.34 & 3.58 & 3.52 & 3.57 & 3.68 \\\midrule
\rowcolor{lightgray}
\texttt{\textbf{DIPLOMAT}} & \textbf{4.23} & \textbf{4.27} & \textbf{4.18} & \textbf{4.54} & \textbf{3.69} & \textbf{4.41} & \textbf{4.60} & \textbf{3.91} & \textbf{4.29} & \textbf{4.16} \\
\bottomrule
\end{tabular}
\end{adjustbox}
\caption{Human evaluation results. All metrics are rated on a scale of 1--5. Results are statistically significant at 5\% significance level based on t-test \cite{welch1947generalization}.}
\label{tab:human_results}
\end{table*}

\section{Experiments}
We compare \texttt{DIPLOMAT} with 
13 baselines: INA \cite{ahmad-etal-2023-ina}, \textsc{genteel-negotiator} \cite{priya2025genteel}, ProCoT (GPT-5-mini) \cite{deng-etal-2023-prompting}, Mistral-7B-Instruct-v0.3-SFT \cite{jiang2023mistral7b}, Llama3.1-8B-SFT \cite{grattafiori2024llama3herdmodels}, Gemma-2-9B-SFT \cite{gemmateam2024gemma2improvingopen}, 
Mistral-7B-Instruct-v0.3-SFT-DPO \cite{rafailov2024directpreferenceoptimizationlanguage}, Mistral-7B-Instruct-v0.3-SFT-ETO \cite{song2024trial}, Mistral-7B-Instruct-v0.3-SFT-DMPO \cite{shi-etal-2024-direct}, and Mistral-7B-Instruct-v0.3-PositiveSFT. For automatic evaluation, we adopt Perplexity (PPL) \cite{brown1992estimate}, BLEU-4 (B-4) \cite{10.3115/1073083.1073135}, BERTScore-F1 (BS-F1) \cite{zhang2019bertscore}, 
Distinct-2 (D-2) \cite{li2015diversity}, and Response Length (R-LEN) to evaluate language quality of responses. To assess responses for task completion, we introduce Negotiation Strategy Correctness (NSC), Persuasion Strategy Correctness (PSC), Politeness Appropriateness (PA), and Agreement Rate (AR). For human evaluation, we employ Fluency (F), Contextual Coherence (CC), and Engagingness (E) to assess language quality of responses. To assess responses for task completeness, we mirror automatic evaluation metrics: NSC, PSC, and PA, report Bargaining Efficacy (BE), Outcome Fairness (OF), and Sociopsychological Closeness (SC) \cite{priya2025genteel}, and introduce Influence Effectiveness (IE). 

Appendix \ref{exp_details} provides `Implementation Details', `Baseline Details', `Evaluation Metrics Details', and `Human Evaluation Process'.  

\section{Results and Analysis}

\subsection{Automatic Evaluation Results} 
Table \ref{tab:automatic_results} presents automatic evaluation results for \texttt{DIPLOMAT} and baselines on the \texttt{PROWESS} dataset. \texttt{DIPLOMAT} consistently achieves superior dialogue quality while maintaining politeness and persuasiveness, reflected in the highest B-4, BS-F1, and D-2 scores, longer responses (R-LEN), effective strategy application (PSC, NSC), high politeness appropriateness (PA), and improved agreement rate (AR).  

Across LLM families, Mistral variants generally improve language quality and task effectiveness more than LLaMA or Gemma models. Compared to the strongest baseline, Mistral-7B-Instruct-v0.3-SFT-ETO, \texttt{DIPLOMAT} achieves gains of 4.1\%, 2.9\%, 1.6\%, 15.1\%, 16.0\%, 13.8\%, 1.1\%, and 6.2\% in B-4, BS-F1, D-2, R-LEN, NSC, PSC, PA, and AR, respectively, indicating linguistically richer, more persuasive, and appropriately polite negotiation responses. Besides, the baselines such as INA and ProCoT (GPT-5-mini) tend to produce shorter, less strategic, or less polite responses, limiting negotiation success. While Mistral-7B variants improve fluency and strategy, they often fail to balance persuasion with politeness. 

Notably, Mistral-7B-Instruct-v0.3-PositiveSFT, which is trained exclusively on preferred dialogue samples, underperforms Mistral-7B-Instruct-v0.3-SFT-DPO across multiple metrics despite both methods having access to the same high-quality preferred data. This suggests that the gains from preference optimization cannot be attributed solely to exposure to higher-quality training samples, but instead arise from the contrastive learning signal provided by negative examples. By explicitly contrasting preferred and non-preferred dialogue spans, preference-based optimization enables the model to learn not only desirable negotiation behaviors, but also which strategic, persuasive, and politeness-related behaviors should be avoided. In contrast, \texttt{DIPLOMAT}'s high performance across all evaluated aspects demonstrates the effectiveness of its dialogue-span-aware preference learning in generating polite, persuasive, and coherent negotiation dialogues that support successful and balanced workplace negotiation outcomes.\\

\subsection{Human Evaluation Results} 
Table \ref{tab:human_results} presents human evaluation results for \texttt{DIPLOMAT} and baseline models on the \texttt{PROWESS} dataset. Overall, \texttt{DIPLOMAT} achieves the highest scores across all metrics, demonstrating its ability to generate fluent (F), coherent (CC), and engaging (E) negotiation dialogues while effectively applying negotiation and persuasion strategies (NSC, PSC) with appropriate politeness (PA). These strengths translate into higher bargaining efficacy (BE), fair outcomes (OF), stronger sociopsychological closeness (SC), and more effective influence (IE), reflecting a holistic negotiation capability. Compared to the strongest baseline, Mistral-7B-Instruct-v0.3-SFT-ETO, \texttt{DIPLOMAT} improves by 2.4\%, 2.2\%, 2.2\%, 30.8\%, 7.6\%, 1.1\%, 19.8\%, 3.7\%, 12.3\%, and 5.6\% in F, CC, E, NSC, PSC, PA, BE, OF, SC, and IE, respectively, reflecting enhanced fluency, coherence, engagement, strategy correctness, politeness, bargaining effectiveness, fairness, social closeness, and influence. These improvements highlight \texttt{DIPLOMAT}'s ability to generate polite, persuasive, and effective workplace negotiation dialogues that sustain natural conversational flow, maintain appropriate strategy use, and consistently support mutually beneficial negotiation outcomes through dialogue-span-aware preference learning.

\subsection{Additional Analyses} We include more analyses: (1) Ablation w.r.t. SFT and Preference Optimization, (2) Analysis on Preference Optimization Granularity, (3) Ablation on Intermediate Blocks in SFT, (4) Sensitivity Analysis on Utility Threshold, (5) Ablation on Utility Components, (6) Effect of Dialogue Span Length, (7) Interlocutor Selection for Preference Data Generation, (8) Out-of-Domain Evaluation, (9) Evaluation Across Negotiation Aspects, and (10) Case Study in Appendix \ref{additional_analysis}.

\section{Conclusion}
We presented \texttt{DIPLOMAT}, a dialogue system designed for polite and persuasive workplace negotiation. For this task, we introduced \texttt{PROWESS}, a workplace negotiation dialogue dataset generated through a multi-agent framework. \texttt{DIPLOMAT} leverages Dialogue-Span-aware DPO (DSA-DPO) objective to improve LLM-based agents in multi-turn negotiation settings, enabling coherent responses that balance politeness and persuasion. Evaluation on \texttt{PROWESS} establish \texttt{DIPLOMAT}'s efficacy in generating polite and persuasive negotiation responses.

\section*{Limitations}
While \texttt{PROWESS} provides a large-scale dataset for studying polite and persuasive workplace negotiations, it is generated using LLM-based multi-agent simulations rather than real human-human interactions. Although we employ diverse scenarios and extensive evaluations to improve realism, synthetic dialogues may not fully capture the variability and subtle social (politeness, persuasiveness, and negotiation) dynamics present in real-world interactions. Furthermore, negotiation strategies, persuasion strategies, and politeness levels are automatically annotated using LLM-based annotation agents. While inter-LLM agreement analysis indicates moderate to fair reliability of these annotations, some ambiguity in interpreting strategies or politeness levels may introduce noise. Finally, the dataset currently focuses on English workplace negotiation scenarios. The proposed generation framework can be extended to other languages and domains with minimal prompt modifications, and exploring multilingual and cross-cultural negotiation settings remains an important direction for future work. 

From the modeling perspective, the proposed \texttt{DIPLOMAT} framework leverages the Dialogue-Span-Aware Direct Preference Optimization (DSA-DPO) objective, which assumes equal-length positive and negative dialogue spans for preference alignment and to simplify optimization process. However, this assumption may limit the granularity of preference alignment, as negative segments can include irrelevant or error-free turns or may fail to capture all problematic dialogue spans. Moreover, the current utility design favors successful agreements and does not explicitly model rational impasse recognition or graceful negotiation termination under hard constraints. Consequently, the model may exhibit a bias toward continued compromise-seeking behavior even in scenarios involving non-negotiable requirements or infeasible negotiation outcomes. Hence, developing methods that support preference alignment over variable-length dialogue spans, while also explicitly modeling dealbreakers and negotiation impasses, remains an open challenge for future work. 

Moreover, while \texttt{DIPLOMAT} and the proposed DSA-DPO objective are designed to be domain-agnostic, our primary evaluation is conducted on workplace negotiation dialogues, as \texttt{PROWESS} is the only existing dataset that jointly models negotiation strategies, persuasion, and politeness in professional interactions. To empirically test generalization beyond this setting, we additionally evaluate \texttt{DIPLOMAT} out-of-domain on the tourism-focused \texttt{NeGoChat} dataset, where it maintains strong negotiation, persuasion, and politeness performance without any domain-specific fine-tuning, demonstrating cross-domain transferability. The remaining limitation is therefore one of breadth rather than of evidence: our out-of-domain analysis covers an additional domain, and a systematic study spanning a wider range of negotiation domains and modalities remains valuable future work. 

\section*{Ethics Statement}
The development of dialogue systems for polite and persuasive workplace negotiations requires careful ethical consideration due to the inherent influence and bargaining dynamics involved in negotiation interactions. This work received approval from our Institutional Review Board (IRB). The proposed system is designed to support constructive and respectful negotiation by encouraging polite communication and responsible persuasion strategies that aim to facilitate mutually beneficial outcomes. Rather than promoting adversarial or manipulative bargaining, the system emphasizes collaborative negotiation and preserves user autonomy, ensuring that individuals remain free to accept or reject any proposed agreement according to their own interests.

Ethical considerations are also important in the construction and use of the dataset. Since the dialogues are generated through LLMs–based simulations, they may reflect biases present in the underlying training data. Such biases could manifest in the representation of workplace roles, negotiation behaviors, or persuasive strategies, potentially reinforcing stereotypes or unfair assumptions. Prior work \cite{taori2023data} has shown that synthetic data generation can amplify biases or create feedback loops if not carefully monitored. To mitigate these risks, the dataset construction process incorporates careful prompt design, human oversight, and evaluation to promote respectful, professional, and contextually appropriate negotiation dialogue.

Furthermore, such dialogue systems raise concerns about potential misuse, such as manipulating individuals or influencing decisions in ways that are not transparent or aligned with users’ interests. Our framework focuses on modeling ethical politeness and persuasion that supports cooperative decision-making and professional communication rather than coercive influence. We encourage future work to further investigate safeguards for responsible polite and persuasive negotiation dialogue generation, including transparency, fairness analysis, and bias mitigation.

Finally, human experts contributed to data validation, filtering, and evaluation, and were compensated in accordance with institutional guidelines. The dataset will be shared for research purposes with appropriate usage guidelines to ensure responsible use. We believe that advancing research on polite, persuasive negotiation dialogue systems can help develop conversational agents that promote constructive workplace communication and responsible decision support.

\section*{Acknowledgement}
The research reported in this paper is an outcome of the project titled \enquote{Conversational Agents with Negotiation and Influencing Ability}, sponsored by Accenture Labs, Banglore, India.


\bibliography{custom}

\clearpage
\onecolumn

{\Large \textbf{Frequently Asked Questions (FAQs)}}

\vspace{6pt}

\noindent \textbf{$\ast$ Why focus on both politeness and persuasion jointly in workplace negotiation?}\\[2pt]
$\twoheadrightarrow$ Workplace negotiation involves balancing task success with maintaining professional relationships. Politeness mitigates interpersonal friction, fosters trust, and prevents escalation, while persuasion helps overcome resistance and align preferences toward mutually beneficial agreements. Treating them in isolation, as prior work has done, overlooks their complementary roles. Hence, jointly modeling both is essential for realistic and effective negotiation dialogue systems.

\vspace{6pt}
\noindent \textbf{$\ast$ Why use synthetic data generated by LLMs instead of real human--human negotiation dialogues?}\\[2pt]
$\twoheadrightarrow$ Collecting large-scale, annotated, multi-turn workplace negotiation dialogues from real human interactions is resource-intensive and raises privacy concerns. Our multi-agent generation framework enables efficient production of high-quality, diverse dialogues with fine-grained annotations for negotiation strategies, persuasion strategies, and politeness levels, while closely emulating natural human negotiation dynamics. Our quantitative validation shows that model-generated utterances are often indistinguishable from human-written text. 

\vspace{6pt}
\noindent \textbf{$\ast$ Why does DSA-DPO operate on dialogue spans rather than individual utterances or full dialogues?}\\[2pt]
$\twoheadrightarrow$ Utterance-level DPO optimizes individual responses but fails to capture evolving negotiation strategies, polite framing, and persuasive justifications that unfold over multiple exchanges. Dialogue-level DPO considers entire conversations but is too coarse-grained, introducing training noise and obscuring the contribution of individual utterances. DSA-DPO addresses both limitations by identifying key dialogue spans that contribute most to negotiation outcomes, reducing noise from non-critical turns while enabling multi-turn preference alignment.

\vspace{6pt}
\noindent \textbf{$\ast$ Why are experiments conducted only on the proposed \texttt{PROWESS} dataset?}\\[2pt]
$\twoheadrightarrow$ Our experiments focus on \texttt{PROWESS} because it is specifically designed to capture \emph{polite and persuasive workplace negotiation}, which is not explicitly modeled in existing negotiation dialogue datasets. Prior datasets primarily focus on bargaining scenarios such as item allocation, e-commerce negotiation, or tourism planning, and therefore lack annotations and conversational structures that reflect professional workplace negotiation dynamics. Since the objectives of this work are to study politeness and persuasion-aware modeling in workplace negotiations, evaluating models on \texttt{PROWESS} provides a controlled and task-relevant benchmark. Nevertheless, the proposed framework and training methodology are general and can be applied to other negotiation dialogue datasets or domains. 

\vspace{6pt}
\noindent \textbf{$\ast$ How are suboptimal utterances identified during preference data construction, given that negotiation errors are inherently ambiguous?}\\[2pt]
$\twoheadrightarrow$ Unlike tasks with objectively defined errors (e.g., math or programming), suboptimal utterances in negotiation are context-dependent and subjective. We use Gemini-2.5-Pro with structured prompts to identify utterances that: (i) are pivotal for advancing the negotiation goal, (ii) exhibit low effectiveness across utility dimensions such as politeness trajectory, strategy alignment, and agreement quality, or (iii) contribute to negotiation breakdown. This utility-driven detection enables us to construct meaningful preference pairs for training.

\vspace{6pt}
\noindent \textbf{$\ast$ What is the utility score and how does it determine dialogue quality for preference learning?}\\[2pt]
$\twoheadrightarrow$ The utility score is a composite metric that evaluates a dialogue across six dimensions: Politeness Trajectory (PT), Persuasion Strategy Alignment (PSA), Negotiation Strategy Alignment (NSA), Mutual Satisfaction (MS), Agreement Quality (AQ), and Negotiation Breakdown (NB). Dialogues with utility scores below a threshold (0.6) are treated as potential negative samples containing suboptimal negotiation behaviors, which are then paired with higher-scoring dialogues to construct dialogue-span-aware preference data for DSA-DPO training.

\twocolumn
\appendix

\section*{Appendix}
\label{sec:appendix}

\section{\texttt{PROWESS} Dataset Details}
\label{dataset_details}

\subsection{Annotation Schema Description}
 Table \ref{negostrategy_def_examples}, Table \ref{perstrategy_def_examples}, and Table \ref{politeness_examples} provide the definitions along with example utterances for different negotiation strategies, persuasion strategies, and politeness levels from the \texttt{PROWESS} dataset. 

\begin{table*}[t]\footnotesize
\centering
\footnotesize
\begin{tabular}{p{0.18\linewidth} | p{0.35\linewidth} | p{0.40\linewidth}}
\toprule
\textbf{Negotiation Strategy} & \textbf{Definition} & \textbf{Example} \\
\midrule

\textbf{Collaborative Style} &
Working jointly to find mutually beneficial outcomes and preserve relationships. &
\textit{Given my experience, I was hoping for a two-year vesting or removal of the one-year cliff.} \\ \hline

\textbf{Active Listening} &
Demonstrating attentiveness and understanding to build rapport and trust. &
\textit{Hi Ben, I'm excited about the possibility of you joining Innovatech. Your expertise in AI is impressive!} \\ \hline

\textbf{Win-Win Framing} &
Framing negotiation as a shared problem to solve for mutual benefit. &
\textit{I see where you're coming from. That would certainly provide immediate benefits, but let's talk about long-term gains.} \\ \hline

\textbf{Principled Negotiation} &
Focusing on mutual interests and objective standards rather than positions. &
\textit{I really value our dialogue, Ben. How about we agree on the standard vesting, with a performance review at 18 months?} \\ \hline

\textbf{Data-Driven Justification} &
Supporting negotiation points with evidence like market benchmarks and past performance. &
\textit{I understand your perspective. May I explain why we maintain this vesting structure? It promotes team stability and fairness.} \\ \hline

\textbf{MESO} &
Proposing multiple offers of equal value to reveal priorities and increase agreement likelihood. &
\textit{I appreciate your dedication, but how about we explore a performance-based bonus structure instead? It could complement the standard vesting.} \\ \hline

\textbf{Anchoring} &
Setting a strong initial offer to influence the negotiation range. &
\textit{Thank you, David. I believe the PMP certification is crucial for my role. I need full funding upfront to accept the offer.} \\ \hline

\textbf{Door-in-the-Face} &
Starting with a larger request to make the actual target seem more acceptable. &
\textit{Absolutely, and I recognize its value. However, our budget constraints are quite tight, given our non-profit status.} \\ \hline

\textbf{Reciprocal Concessions} &
Offering small concessions to encourage reciprocation from the other party. &
\textit{That sounds reasonable, but I still believe an even earlier vesting period could align our interests better.} \\ \hline

\textbf{Credibility Assertion} &
Building trust by reinforcing personal or organizational credibility during negotiations. &
\textit{Ben, I appreciate your experience. However, our standard four-year vesting with a one-year cliff ensures fairness across our team.} \\ \hline

\textbf{No Strategy} &
Indicates the absence of any explicit negotiation strategy. &
\textit{Thank you, Anya. I'm enthusiastic about the role, but I need to discuss the vesting schedule.} \\

\bottomrule
\end{tabular}
\caption{Definitions and example utterances of negotiation strategies in the \texttt{PROWESS} dataset.}
\label{negostrategy_def_examples}
\end{table*}

\begin{table*}[t]\footnotesize
\centering
\footnotesize
\begin{tabular}{p{0.18\linewidth} | p{0.35\linewidth} | p{0.40\linewidth}}
\toprule
\textbf{Persuasion Strategy} & \textbf{Definition} & \textbf{Example} \\
\midrule

\textbf{Rapport Building} &
Establishing trust and a positive relationship to enhance receptiveness. &
\textit{Hi Ben, I'm excited about the possibility of you joining Innovatech. Your expertise in AI is impressive!} \\ \hline

\textbf{Concern Addressing} &
Actively listening to and resolving objections or hesitations. &
\textit{I understand that the four-year vesting with a one-year cliff may feel lengthy, but let's discuss this further.} \\ \hline

\textbf{Emotional Appeal} &
Connecting through emotions such as empathy, excitement, or urgency. &
\textit{I understand the importance of long-term commitment, but I believe my skills warrant quicker rewards.} \\ \hline

\textbf{Credibility \& Confidence} &
Demonstrating expertise and confidence to increase trust and influence. &
\textit{Hi Ben, I'm thrilled about the possibility of you joining Innovatech. Let's discuss your concerns regarding the equity vesting schedule.} \\ \hline

\textbf{Data-Driven Persuasion} &
Using evidence, facts, and benchmarks to strengthen arguments. &
\textit{But wouldn't an accelerated vesting schedule incentivize me even more to contribute significantly right from the start?} \\ \hline

\textbf{Problem-Solving Focus} &
Presenting ideas as solutions that address mutual challenges. &
\textit{I understand your perspective. May I explain why we maintain this vesting structure? It promotes team stability and fairness.} \\ \hline

\textbf{Self-Interest Appeal} &
Framing arguments around how the outcome directly benefits the other party. &
\textit{That sounds reasonable, but I still believe an even earlier vesting period could align our interests better.} \\ \hline

\textbf{Value Alignment} &
Linking your proposal to the other party's core values and principles. &
\textit{I assure you that we want to reward your hard work. Let's finalize a review after the first 18 months.} \\ \hline

\textbf{Reputation Highlighting} &
Leveraging past achievements or organizational standing to reinforce trustworthiness. &
\textit{I can assure you, our track record supports rewarding outstanding performance with recognition and opportunities for early vesting.} \\ \hline

\textbf{Future Vision Alignment} &
Connecting the proposal with shared long-term goals and aspirations. &
\textit{I really value our dialogue, Ben. How about we agree on the standard vesting, with a performance review at 18 months?} \\ \hline

\textbf{No Strategy} &
Indicates the absence of any explicit persuasion strategy. &
\textit{Absolutely, and I see the value it brings. However, upfront funding is challenging for our tight budget.} \\

\bottomrule
\end{tabular}
\caption{Definitions and example utterances of persuasion strategies in the \texttt{PROWESS} dataset.}
\label{perstrategy_def_examples}
\end{table*}

\begin{table*}[hbt!]\footnotesize
\centering
\small
\begin{tabular}{p{0.18\linewidth} | p{0.35\linewidth} | p{0.40\linewidth}}
\toprule
\textbf{Politeness Level} & \textbf{Definition} & \textbf{Example} \\ 
\midrule
\textbf{High\_Polite} & Highly courteous language emphasizing warmth and rapport. & I’m glad to hear that, Maria. Your enthusiasm is contagious, and we’re excited about your potential contributions. \\ \hline

\textbf{Moderate\_Polite} & Respectful and cooperative tone maintaining professional balance. & I appreciate your concern, Ben. Our past hires have thrived under this structure, which builds trust and shared goals. \\ \hline

\textbf{Low\_Polite} & Direct and assertive language with minimal politeness markers. & A partial stipend is better than nothing, but I need assurance of full funding eventually. \\

\bottomrule
\end{tabular}
\caption{Definitions and example utterances of different politeness levels in the \texttt{PROWESS} dataset.}
\label{politeness_examples}
\end{table*}

\subsection{Details of Dataset Quality Assessment Criteria}
\begin{enumerate}[leftmargin=1.2em,itemsep=0pt, parsep=0pt, topsep=0pt, partopsep=0pt]
\setlength{\itemsep}{0pt}
\setlength{\parskip}{0pt}
\setlength{\parsep}{0pt}
    \item \textbf{Scenario Consistency (SC)}: Does the dialogue stay aligned with the given negotiation scenario?
    \item \textbf{Negotiation Strategy Correctness (NSC)}: Does each utterance employ the negotiation strategy correctly, given the speaker's role (employer or candidate) and the current dialogue context?
    \item \textbf{Persuasion Strategy Correctness (PSC)}: Does each utterance employ the persuasion strategy correctly, given the speaker's role (employer or candidate) and the current dialogue context?
    \item \textbf{Politeness Appropriateness (PA)}: Does each utterance demonstrate suitable level of politeness given the context, role, and interaction dynamics?
    \item \textbf{Fairness (F)}: Does the outcome of the dialogue reflect a balanced, equitable, and mutually beneficial resolution?
    \item \textbf{Coherence (C)}: Is the dialogue logically structured, with smooth progression and clear connections between utterances?
    \item \textbf{Naturalness (N)}: Does the conversation resemble human negotiation in language, tone, and flow?
    \item \textbf{Engagingness (E)}: Is the dialogue interesting, interactive, and rich enough to sustain the user's attention throughout the negotiation?
\end{enumerate}

\subsection{Additional Analysis of \texttt{PROWESS} Dataset}
\noindent \textbf{Human-LLM Differentiation Assessment.} To measure linguistic naturalness in \texttt{PROWESS}, we evaluate whether humans can distinguish model-generated utterances from human-rewritten counterparts. We randomly select 600 utterances from 240 dialogues, ensuring balanced representation from both employer and candidate roles and coverage across different stages of each dialogue. Two native English speakers from our research group manually rewrite each sampled utterance to preserve meaning while improving clarity, fluency, and grammatical correctness, resulting in paired examples of model-generated and human-edited utterances. The same three human evaluators involved in the data quality assessment independently identify the model-generated utterances in each pair. Across the 600 utterance pairs, all evaluators correctly identified the model-generated sentence in 276 cases (46\%), disagreed in 184 pairs (30.7\%), and incorrectly selected the human-edited sentence in 140 pairs (23.3\%), with individual evaluator accuracies of 0.731, 0.747, and 0.722. These results indicate that model-generated utterances in \texttt{PROWESS} are often indistinguishable from human-written text, demonstrating high linguistic naturalness and realism.

\noindent \textbf{Annotation Reliability.} We further evaluate the reliability of automatically generated annotations using a multi-LLM assessment framework. From a random sample of 240 dialogues, each utterance is independently annotated by three state-of-the-art LLMs: \texttt{GPT-4o-mini} (via the Dialogue Annotation Agent), \texttt{Gemini-2.5-Pro}, and \texttt{LLaMA-3-8B}, for negotiation strategies, persuasion strategies, and politeness levels. We use the same prompt template as the Dialogue Annotation Agent to ensure annotation consistency. The inter-LLM Kappa agreement scores ($\kappa$) \cite{mchugh2012interrater} are found to be 0.724 for negotiation strategies, 0.742 for persuasion strategies, and 0.803 for politeness levels. These substantial agreement scores indicate that the strategy and politeness taxonomies are well-defined, enabling consistent annotation while reflecting the inherent complexity and subtlety of polite, persuasive and negotiation behaviors.

\begin{figure}[!h]
    \centering
    \includegraphics[width=\linewidth,trim={0 0 0 0},scale = 1.0]{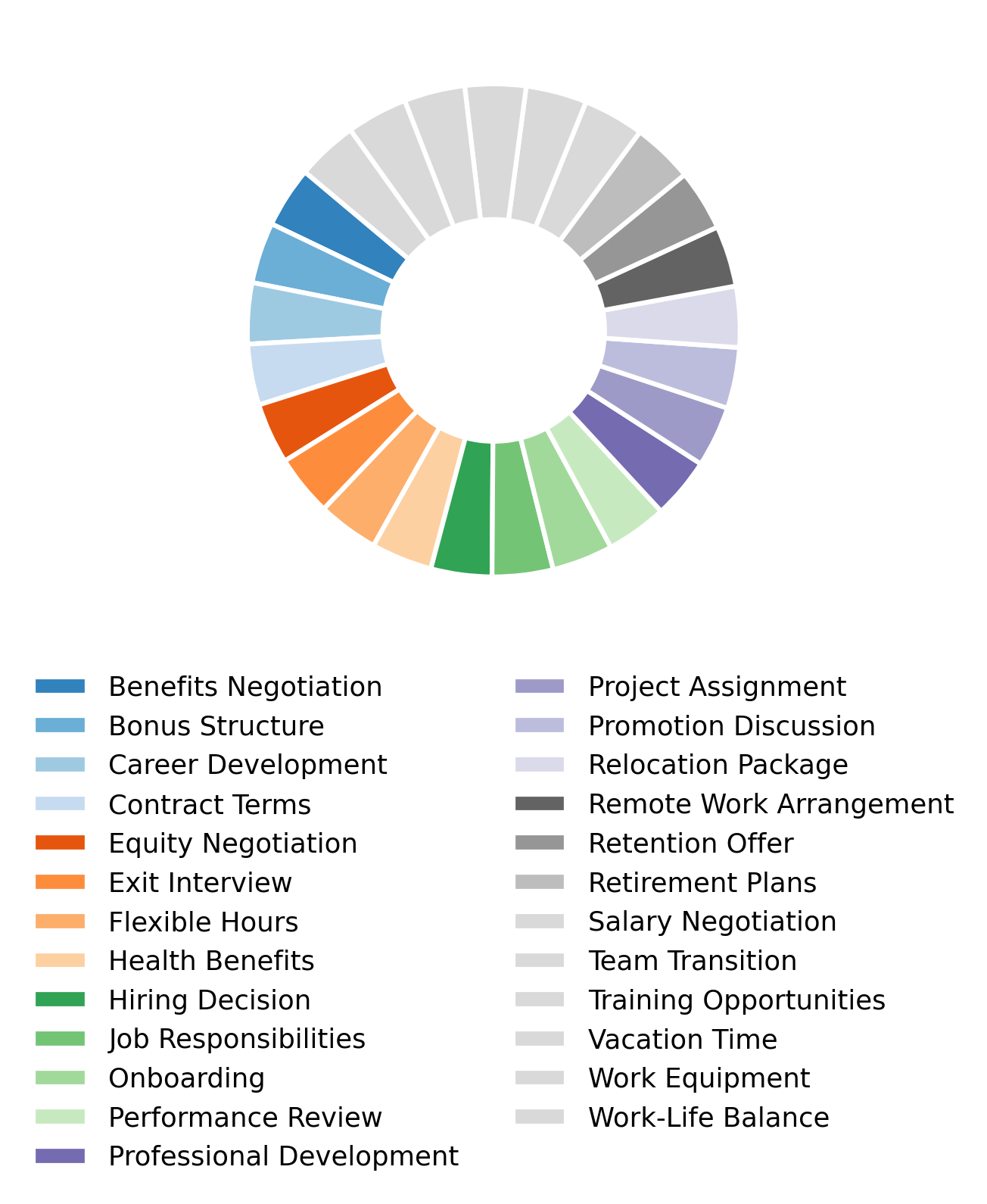}
    \caption{Negotiation aspects-wise dialogue distribution in the \texttt{PROWESS} dataset.}
    \label{negotiation_aspects_pie_chart}
\end{figure}

\begin{figure*}[!h]
    \centering
    \includegraphics[width=\linewidth,trim={0 0 0 0},scale = 1.0]{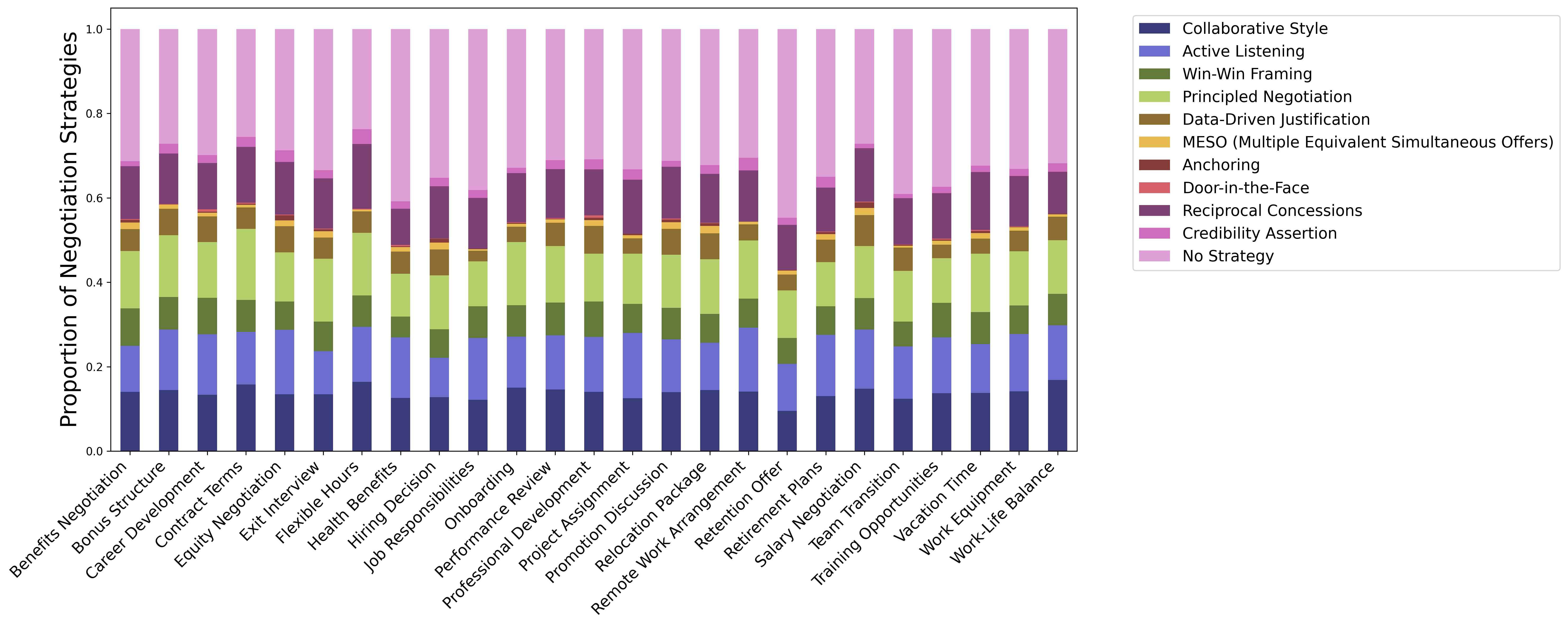}
    \caption{Distribution of negotiation strategies across different negotiation aspects in the \texttt{PROWESS} dataset.}
    \label{nego_dist}
\end{figure*}

\begin{figure*}[!h]
    \centering
    \includegraphics[width=\linewidth,trim={0 0 0 0},scale = 1.0]{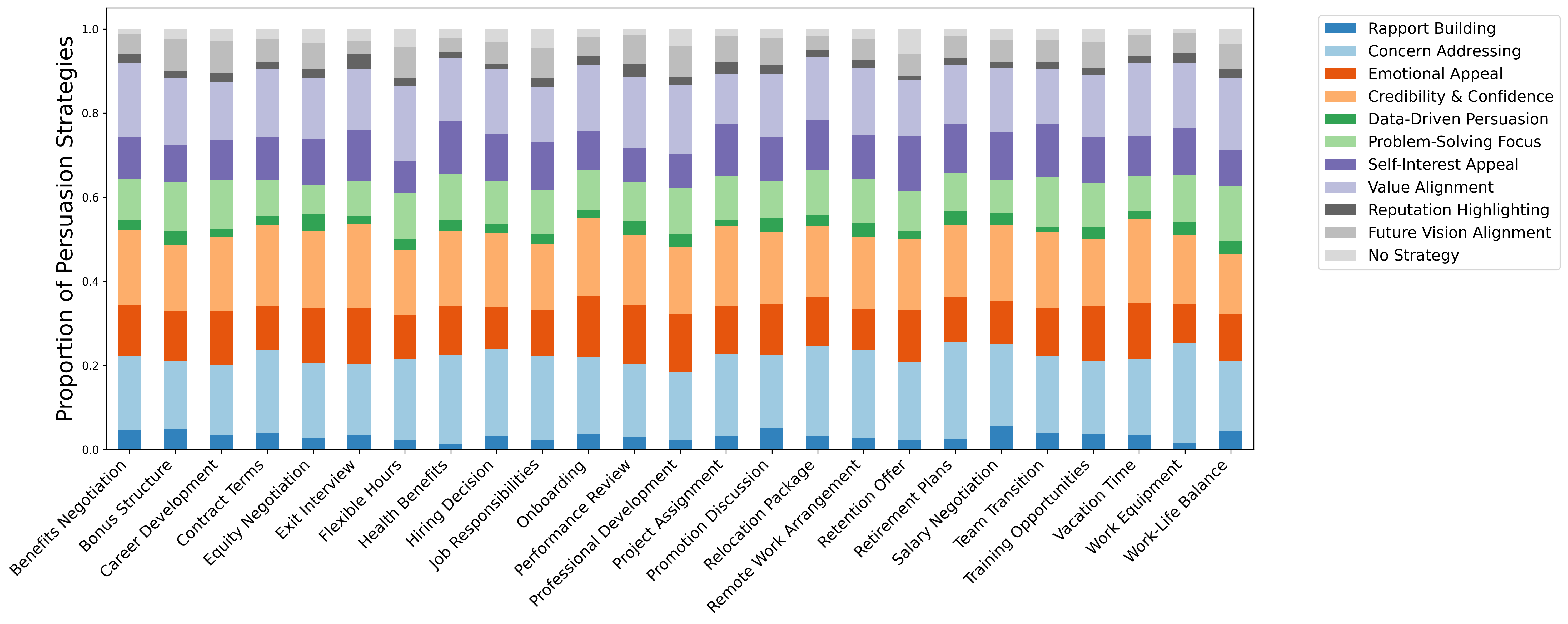}
    \caption{Distribution of persuasion strategies across different negotiation aspects in the \texttt{PROWESS} dataset.}
    \label{persu_dist}
\end{figure*}

\begin{figure*}[!h]
    \centering
    \includegraphics[width=\linewidth,trim={0 0 0 0},scale = 1.0]{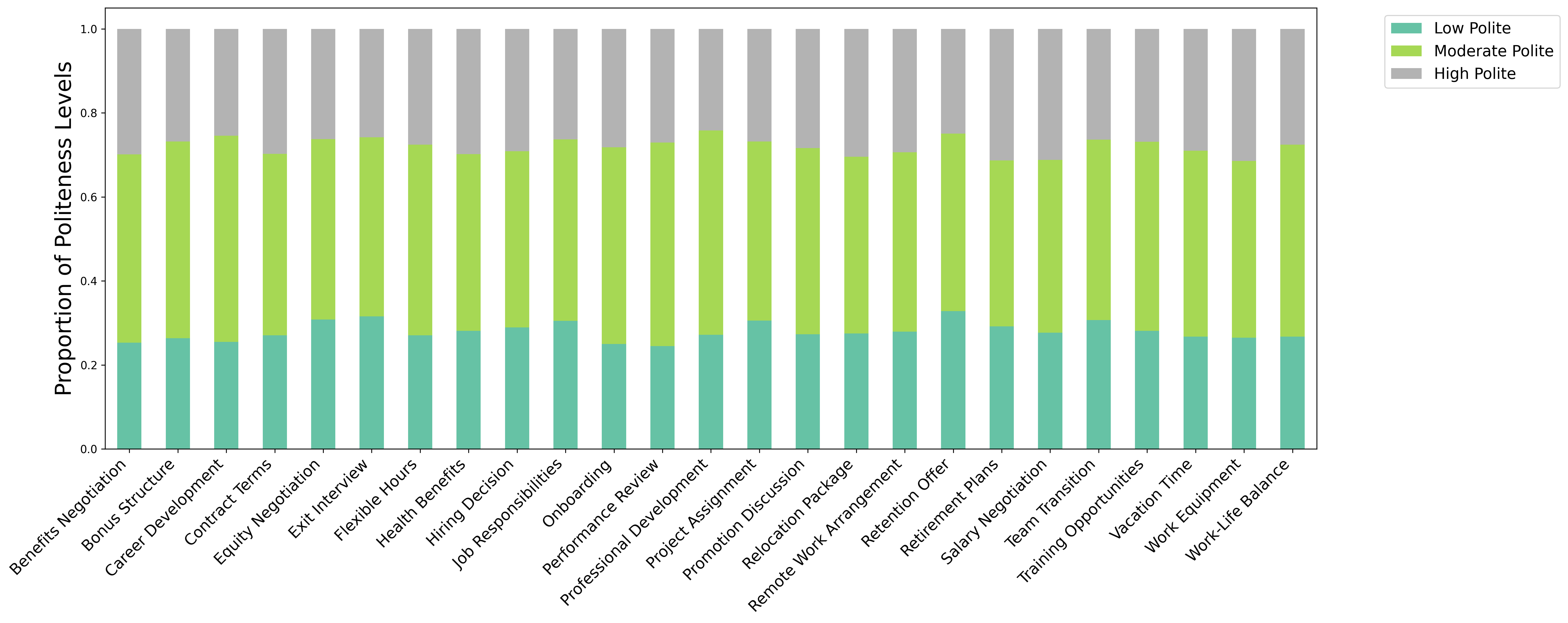}
    \caption{Distribution of politeness levels across different negotiation aspects in the \texttt{PROWESS} dataset.}
    \label{polite_dist}
\end{figure*}

\noindent \textbf{Dataset Diversity.} As illustrated in Figure~\ref{negotiation_aspects_pie_chart}, the \texttt{PROWESS} dataset exhibits balanced coverage across negotiation aspects, ensuring uniform representation and preventing topic skew. This balance enables robust evaluation of dialogue models across diverse negotiation scenarios. The distributions of negotiation strategies, persuasion strategies, and politeness levels are shown in Figures~\ref{nego_dist}, \ref{persu_dist}, and \ref{polite_dist}, respectively. Overall, negotiation strategies such as a collaborative style, win-win framing, and active listening are more common, reflecting the objective of reaching mutually beneficial agreements. However, strategy usage varies across negotiation aspects, indicating context-sensitive negotiation behaviors. For instance, anchoring and MESO strategies are more common in compensation-related discussions, where concrete offers and trade-offs are central, whereas principled negotiation and credibility assertions are more common in role- or responsibility-focused dialogues that emphasize fairness and trust. A similar trend is observed for persuasion strategies: logical and outcome-oriented appeals are generally prevalent, while emotional appeals occur more often in personal discussions such as career development, and credibility or data-driven justifications are more prominent in finance- or policy-oriented negotiations. The distribution of politeness levels further shows a balanced mix of low, medium, and high politeness across aspects, with higher politeness typically appearing in sensitive or relationship-oriented contexts and more direct language used in transactional discussions. Overall, these patterns suggest that the dataset captures diverse and contextually adaptive polite and persuasive negotiation behaviors throughout interactions.

\noindent \textbf{Comparison with Existing Negotiation Dialogue Datasets.} We present a comparison of the proposed \texttt{{PROWESS}} dataset with existing publicly available negotiation dialogue datasets in Table~\ref{compare_datasets}. While prior datasets primarily focus on task-oriented bargaining scenarios such as item allocation, e-commerce negotiation, or tourism planning, \texttt{{PROWESS}} introduces a dataset centered on \emph{polite and persuasive workplace negotiation}, capturing interactions that emphasize professionalism, rapport building, and constructive influence through politeness and persuasion modeling in organizational contexts. Furthermore, \texttt{{PROWESS}} exhibits an average dialogue length of 21.58 utterances, which is substantially longer than most existing negotiation datasets, reflecting richer multi-turn interactions and more realistic negotiation dynamics. With 2,400 dialogues and over 51K utterances, the dataset provides a substantial collection of negotiation conversations within workplace settings, enabling the study of negotiation techniques, persuasive communication, and politeness-aware dialogue modeling. These characteristics make \texttt{{PROWESS}} a valuable benchmark for developing and evaluating dialogue agents capable of conducting polite and persuasive negotiations in workplace environments.

\begin{table*}[t]
\centering
\small
\begin{adjustbox}{max width=\linewidth}
\begin{tabular}{l l c r r r}
\toprule
\textbf{Dataset} & \textbf{Domain} & \textbf{Setting} & \textbf{\#Dialogues} & \textbf{\#Utterances} & \textbf{Avg. Dialogue Length} \\
\midrule
STAC \cite{asher2016discourse} & Strategy Games & I & 1,081 & 9,188 & 8.5 \\
DealorNoDeal \cite{lewis-etal-2017-deal} & Item Assignment & I & 5,808 & 38,332 & 6.6 \\
CraigslistBargain \cite{he-etal-2018-decoupling} & E-commerce & D & 6,682 & 61,474 & 9.2 \\
NegoCoach \cite{zhou2019dynamic} & Product Bargaining & D & 300 & -- & -- \\
Anti-Scam \cite{li2020end} & E-commerce & D & 220 & 2,739 & 12.45 \\
CaSiNo \cite{chawla-etal-2021-casino} & Item Assignment & I & 1,030 & 11,948 & 11.6 \\
JobInterview \cite{yamaguchi-etal-2021-dialogue} & Job Interview & I & 2,639 & 33,515 & 12.7 \\
IND \cite{ahmad-etal-2023-ina} & E-commerce & I & 4,163 & 57,393 & 13.79 \\
DEAL \cite{priya2024trip} & Tourism & I & 1,291 & 18,932 & 14.66 \\
PACT \cite{priya2025we} & Tourism & I & 8,687 & 186,241 & 21.43 \\
NEGOCHAT \cite{priya2025genteel} & Tourism & I & 1,242 & 18,638 & 15.01 \\\bottomrule
\texttt{PROWESS} (Proposed) & Workplace & I & 2,400 & 51,796 & 21.58 \\
\bottomrule
\end{tabular}
\end{adjustbox}
\caption{Comparison of negotiation dialogue datasets. I: integrative negotiation, D: distributive negotiation.}
\label{compare_datasets}
\end{table*}

\section{Evaluation of Gemini-2.5-Pro in Preference Data Construction}
\label{app_pref_data}
To assess the reliability of the LLM-based pipeline used for preference data construction, we conduct a manual evaluation of the outputs generated by Gemini-2.5-Pro. The evaluation of negotiation dialogues is inherently challenging due to the subjective nature of politeness, persuasion, and negotiation. As a result, strict notions of alignment are often difficult to establish. To account for this ambiguity, we adopt three evaluation categories: \textit{Aligned}, \textit{Plausible}, and \textit{Misaligned}. A prediction is labeled `\textit{Aligned}' when the model’s decision is consistent with human judgment, `\textit{Plausible}' when the model’s decision appears reasonable but may not represent the most appropriate choice, and `\textit{Misaligned}' when the model’s output clearly contradicts human judgment.

For this analysis, we randomly sample 140 preference pairs from the constructed preference dataset. The same three human evaluators involved in the dataset quality assessment independently assess Gemini-2.5-Pro’s performance across three stages of the pipeline: utility evaluation, suboptimal utterance detection, and dialogue-span extraction. The average evaluation results are reported in Table~\ref{gemini_pref_evaluation}. The evaluators observe that most \textit{Plausible} cases arise when Gemini-2.5-Pro produces outputs that are reasonable in the given conversational context, but determining whether they represent the most informative or strategically appropriate choice remains difficult. Across all three components, \textit{Misaligned} cases are relatively rare, indicating that the model seldom produces clearly incorrect judgments. Overall, these results suggest that Gemini-2.5-Pro provides effective outputs for utility evaluation, suboptimal utterance detection, and dialogue-span extraction within the preference data construction pipeline.

\begin{table*}[t]
\centering
\small
\begin{tabular}{p{0.6\linewidth}ccc}
\toprule
\textbf{Component} & \textbf{Aligned} & \textbf{Plausible} & \textbf{Misaligned} \\
\midrule
Utility Evaluation & 100.9 & 31.4 & 7.7 \\
Suboptimal Utterance Detection & 98.6 & 33.1 & 8.3 \\
Dialogue-Span Extraction & 101.2 & 30.5 & 8.3 \\
\bottomrule
\end{tabular}
\caption{Results of manual evaluation of Gemini-2.5-Pro's performance in preference data construction. The values represent the average number of samples assigned to each evaluation category across the three evaluators.}
\label{gemini_pref_evaluation}
\end{table*}

\section{Experiment Details}
\label{exp_details}

\subsection{Implementation Details}
All models are implemented in the PyTorch framework\footnote{\url{https://pytorch.org/}}, and we use transformer-based architectures from the Hugging Face library \cite{wolf2019huggingface}. For OpenAI-based experiments, we access \texttt{gpt-4o-mini} and \texttt{gpt-5-mini} through the API. During the SFT stage, we employ QLoRA with rank $r{=}8$ and 4-bit quantization. The optimization is performed using the AdamW optimizer \citep{kingma2014adam} with a learning rate of $1{\times}10^{-5}$, a 5\% warm-up ratio, and a cosine decay learning rate schedule. The model is trained for 3 epochs with a batch size of 4 and a dropout rate of 0.2. For response generation, we use top-$k{=}30$ and top-$p{=}0.9$ sampling \cite{holtzman2019curious}, a temperature of 0.7, and a repetition penalty of 1.03. In the DSA-DPO optimization stage, $\beta$ is set to 0.1, and the learning rate is set to $5{\times}10^{-6}$ with a cosine decay schedule and no warm-up phase. DSA-DPO training is conducted for 3 epochs with a batch size of 4. All experiments are performed on an NVIDIA A100 80GB GPU using CUDA 12.8.

\subsection{Baseline Details}
\begin{enumerate}[leftmargin=1.2em,itemsep=2pt, parsep=0pt, topsep=2pt, partopsep=0pt]
\item \textbf{INA} \cite{ahmad-etal-2023-ina}: A GPT-2-based dialogue agent trained using reinforcement learning with task-relevance rewards to dynamically adjust prices and manage bundle deals for effective integrative negotiations.

\item \textbf{\textsc{genteel-negotiator}} \cite{priya2025genteel}: A mixture-of-expert-based reinforcement learning framework for polite negotiation dialogue that integrates negotiation, politeness, and keyterm experts through multi-task learning and RL-based rewards.

\item \textbf{ProCoT (GPT-5-mini)} \cite{deng-etal-2023-prompting}: Prompts GPT-5-mini using chain-of-thought to generate a descriptive analysis for proactively planning the action for the next turn.

\item \textbf{Mistral-7B-Instruct-v0.3-SFT} \cite{jiang2023mistral7b}: Mistral-7B-Instruct-v0.3 fine-tuned on the \texttt{PROWESS} training data in a supervised setting.

\item \textbf{Llama3.1-8B-SFT} \cite{grattafiori2024llama3herdmodels}: Llama-3.1-8B fine-tuned on the \texttt{PROWESS} training data in a supervised setting.

\item \textbf{Gemma-2-9B-SFT} \cite{gemmateam2024gemma2improvingopen}: Gemma-2-9B fine-tuned on the \texttt{PROWESS} training data in a supervised setting.

\item \textbf{Mistral-7B-Instruct-v0.3-SFT-DPO} \cite{rafailov2024directpreferenceoptimizationlanguage}: Mistral-7B-Instruct-v0.3 first fine-tuned with SFT on the \texttt{PROWESS} training data, then further aligned using DPO, which directly optimizes the policy to prefer chosen over rejected responses using a pairwise preference loss.

\item \textbf{Mistral-7B-Instruct-v0.3-SFT-ETO} \cite{song2024trial}: Mistral-7B-Instruct-v0.3 first fine-tuned with SFT on the \texttt{PROWESS} training data, then further aligned using Exploration-based Trajectory Optimization (ETO), which employs trial-and-error exploration to construct contrastive trajectory pairs for trajectory-level optimization.

\item \textbf{Mistral-7B-Instruct-v0.3-SFT-DMPO} \cite{shi-etal-2024-direct}: Mistral-7B-Instruct-v0.3 first fine-tuned with SFT on the \texttt{PROWESS} training data, then further aligned using Direct Multi-turn Preference Optimization (DMPO), which extends DPO to multi-turn dialogue settings by incorporating turn-level credit assignment.

\item \textbf{Mistral-7B-Instruct-v0.3-PositiveSFT}: Mistral-7B-Instruct-v0.3 fine-tuned using supervised fine-tuning exclusively on the positive (chosen) dialogue samples from the preference dataset.
\end{enumerate}

\subsection{Evaluation Metrics Details}
\paragraph{Automatic Evaluation Metrics.} Perplexity (PPL) \cite{brown1992estimate} evaluates the model's ability to predict the next token in a sequence, reflecting the fluency of generated responses. BLEU-4 (B-4) \cite{10.3115/1073083.1073135} measures the degree of 4-gram overlap between generated outputs and reference responses. Distinct-2 (D-2) \cite{li2015diversity} quantifies response diversity by calculating the ratio of unique bigrams within the generated text. BERTScore-F1 (BS-F1) \cite{zhang2019bertscore}\footnote{BERTScore: \url{https://huggingface.co/spaces/evaluate-metric/bertscore}} evaluates semantic similarity between generated and reference responses using contextual embeddings from BERT \cite{devlin2018bert}. Negotiation Strategy Correctness (NSC) measures whether the generated response appropriately applies the intended negotiation strategy. Persuasion Strategy Correctness (PSC) evaluates whether the response correctly employs the intended persuasion strategy. Politeness Appropriateness (PA) assesses whether the generated response maintains an appropriate level of politeness. Agreement Rate (AR) measures the percentage of dialogues that result in a successful negotiation agreement between participants.
\[
\text{NSC} = 
\mathbb{E}_{(\mathcal{C}_t,e_t)}
\left[
\mathbf{1}
\left\{
f_{\text{neg}}(e_t)
=
f_{\text{neg}}(\hat{e}_t)
\right\}
\right]
\]
\[
\text{PSC} = 
\mathbb{E}_{(\mathcal{C}_t,e_t)}
\left[
\mathbf{1}
\left\{
f_{\text{per}}(e_t)
=
f_{\text{per}}(\hat{e}_t)
\right\}
\right]
\]
\[
\text{PA} = 
\mathbb{E}_{(\mathcal{C}_t,e_t)}
\left[
\mathbf{1}
\left\{
f_{\text{pol}}(e_t)
=
f_{\text{pol}}(\hat{e}_t)
\right\}
\right]
\]
\[
\text{AR} = 
\frac{1}{N}
\sum_{i=1}^{N}
\mathbf{1}
\left\{
f_{\text{agr}}(d_i)=1
\right\}
\]

where \(\mathcal{C}_t\) denotes the conversational context up to turn \(t\), 
\(e_t\) and \(\hat{e}_t\) denote the ground-truth and generated employer responses at turn \(t\), respectively, and \(f_{\text{neg}}(\cdot)\), \(f_{\text{per}}(\cdot)\), \(f_{\text{pol}}(\cdot)\), and \(f_{\text{agr}}(\cdot)\) denote the negotiation strategy, persuasion strategy, politeness, and agreement classifiers, respectively. \(f_{\text{neg}}(\cdot)\), \(f_{\text{per}}(\cdot)\), \(f_{\text{pol}}(\cdot)\), and \(f_{\text{agr}}(\cdot)\) achieve accuracies of \(84.3\%\), \(86.1\%\), \(89.2\%\), and \(91.4\%\), respectively, and macro-F1 scores of \(81.7\%\), \(83.4\%\), \(86.8\%\), and \(90.1\%\), respectively, on the \texttt{PROWESS} test set.

\paragraph{Human Evaluation Metrics.} Fluency (F) measures the grammatical correctness and overall linguistic quality of the generated responses. Contextual Coherence (CC) evaluates whether the responses are logically connected and relevant to the ongoing dialogue context. Engagingness (E) reflects how interesting and interactive the conversation appears to the user. Bargaining Efficacy (BE) assesses the system’s capability to negotiate effectively by presenting arguments, incentives, or trade-offs that may influence the counterpart’s decisions. Outcome Fairness (OF) assesses whether the final negotiation outcome reflects a balanced, mutually acceptable agreement between participants. In addition, we evaluate NSC, PSC, and PA. While these metrics correspond to their automatic evaluation counterparts, they are assessed by human evaluators to provide a more reliable judgment of strategy usage and emotional appropriateness in the dialogue. Sociopsychological Closeness (SC) measures the extent to which polite responses foster perceived social and emotional proximity between participants. Influence Effectiveness (IE) evaluates how effectively persuasive responses shape the counterpart’s attitude, preferences, or decisions during the negotiation.

\paragraph{Human Evaluation Process.} The human evaluation is performed by three independent evaluators\footnote{These evaluators are different from those involved in dataset preparation and are compensated according to institutional guidelines.}. Two evaluators hold Ph.D. degrees in Linguistics, while the third holds a postgraduate degree in Computer Science. All evaluators have prior experience with dialogue evaluation tasks. Before the evaluation begins, they are provided with instructions and background information regarding the negotiation scenarios. Each evaluator conducts multi-turn negotiation interactions with the system for a given scenario and completes 30 dialogues using different response sets, resulting in a total of 90 human-evaluated negotiation dialogues. After completing each interaction, the evaluators rate the dialogues on F, CC, E, NSC, PSC, PA, BE, OF, SC, and IE using a 1–5 scale (low to high). To assess inter-evaluator agreement, we compute Fleiss’ Kappa ($\kappa$) \cite{mchugh2012interrater}, and obtain $\kappa = 0.76$ (F), $0.74$ (CC), $0.75$ (E), $0.80$ (NSC), $0.81$ (PSC), $0.83$ (PA), $0.74$ (BE), $0.73$ (OF), $0.75$ (SC), and $0.77$ (IE), indicating substantial agreement among the evaluators.

\section{Additional Analysis}
\label{additional_analysis}

\subsection{Ablation w.r.t. SFT and Preference Optimization}
We perform an ablation study to evaluate the contributions of the core learning objectives in \texttt{DIPLOMAT}, namely supervised fine-tuning ($\mathcal{L}_{\text{SFT}}$) and dialogue-span-aware direct preference optimization ($\mathcal{L}_{\text{DSA-DPO}}$). Table~\ref{tab:ablation_objectives} reports results for ablated variants where one or both objectives are removed. The removal of $\mathcal{L}_{\text{SFT}}$ (-SFT) leads to substantial degradation across NSC, PSC, PA, and AR, highlighting the critical role of supervised fine-tuning in equipping the model with foundational negotiation skills and context-aware polite and persuasive behaviors. Further, omitting $\mathcal{L}_{\text{DSA-DPO}}$ (-DSA-DPO) also lowers performance, particularly in NSC and AR, indicating that preference-based optimization is essential for refining negotiation strategies over multi-turn dialogue spans. The joint removal of both objectives (-(SFT + DSA-DPO)) results in the sharpest decline, with PPL increasing and overall task completion metrics dropping considerably. This pattern confirms that SFT and dialogue-span-aware preference optimization are complementary: SFT provides foundational behavior modeling, while DSA-DPO further aligns the model with polite and persuasive negotiation outcomes, and both objectives are necessary to achieve the high performance observed in \texttt{DIPLOMAT}.

\begin{table*}[t]
\centering
\small
\begin{adjustbox}{max width=\linewidth}
\begin{tabular}{lccccccccc}
\toprule
\textbf{Models} & \textbf{PPL $\downarrow$} & \textbf{B-4 $\uparrow$} & \textbf{BS-F1 $\uparrow$} & \textbf{D-2 $\uparrow$} & \textbf{R-LEN $\uparrow$} & \textbf{NSC $\uparrow$} & \textbf{PSC $\uparrow$} & \textbf{PA $\uparrow$} & \textbf{AR (\%) $\uparrow$} \\
\midrule
\rowcolor{lightgray}
\textbf{\texttt{DIPLOMAT}} & \textbf{8.6} & \textbf{0.452} & \textbf{0.630} & \textbf{0.635} & \textbf{19.8} & \textbf{0.724} & \textbf{0.628} & \textbf{0.882} & \textbf{82.3} \\
\bottomrule
\hspace{5mm}- SFT & 12.5 & 0.355 & 0.598 & 0.580 & 18.5 & 0.605 & 0.540 & 0.870 & 73.0 \\
\hspace{5mm}- DSA-DPO & 9.1 & 0.373 & 0.601 & 0.594 & 16.3 & 0.558 & 0.490 & 0.863 & 68.6 \\
\hspace{5mm}- (SFT + DSA-DPO) & 22.4 & 0.229 & 0.569 & 0.352 & 14.2 & 0.412 & 0.372 & 0.841 & 60.2 \\
 \bottomrule
\end{tabular}
\end{adjustbox}
\caption{Ablation of SFT and preference optimization objective in the \texttt{DIPLOMAT}. \textbf{-} signifies the removal of the component.}
\label{tab:ablation_objectives}
\end{table*}


\subsection{Analysis on Preference Optimization Granularity}
To assess the impact of preference optimization granularity in \texttt{DIPLOMAT}, we compare three variants: (i) \emph{utterance-level DPO}, where preferences are applied to individual utterances, (ii) \emph{dialogue-level DPO}, where the entire positive and negative dialogue samples are used during preference optimization, and (iii) our \emph{dialogue-span-aware DPO} (the proposed \texttt{DIPLOMAT} system). Table~\ref{tab:granularity_ablation} shows the results across these settings. The utterance-level DPO improves over SFT alone but cannot fully capture multi-turn negotiation dynamics, limiting the model’s ability to consistently mirror politeness cues and adapt persuasive strategies over the course of a dialogue. As a result, gains in metrics such as NSC, PSC, PA, and AR are moderate. The dialogue-level DPO better models overall negotiation outcomes, leading to higher agreement rates, but lacks fine-grained supervision, which leads to less precise adaptation of politeness, and persuasion and negotiation strategy selection, sometimes generating generic or over-conceding responses. In contrast, dialogue-span-aware DPO (\texttt{DIPLOMAT}) consistently outperforms both alternatives, demonstrating that optimizing over targeted dialogue spans provides clearer and more actionable preference signals. This allows the model to align responses with strategic negotiation behaviors, maintain dynamic politeness trajectories, and adjust persuasion strategies in context-sensitive ways across multiple turns, resulting in more effective and nuanced negotiation outcomes.

\begin{table*}[t]
\centering
\small
\begin{adjustbox}{max width=\linewidth}
\begin{tabular}{lccccccccc}
\toprule
\textbf{Models} & \textbf{PPL $\downarrow$} & \textbf{B-4 $\uparrow$} & \textbf{BS-F1 $\uparrow$} & \textbf{D-2 $\uparrow$} & \textbf{R-LEN $\uparrow$} & \textbf{NSC $\uparrow$} & \textbf{PSC $\uparrow$} & \textbf{PA $\uparrow$} & \textbf{AR (\%) $\uparrow$} \\
\midrule
Mistral-7B-Instruct-v0.3-SFT & 9.1 & 0.373 & 0.601 & 0.594 & 16.3 & 0.558 & 0.490 & 0.863 & 68.6 \\
\hspace{5mm}+ Utterance-level DPO & 8.8 & 0.407 & 0.606 & 0.615 & 18.5 & 0.598 & 0.528 & 0.869 & 73.4 \\
\hspace{5mm}+ Dialogue-level DPO & 8.7 & 0.410 & 0.628 & 0.615 & 18.9 & 0.612 & 0.540 & 0.875 & 74.2 \\\midrule
\rowcolor{lightgray}
\hspace{5mm}+ Dialogue-Span-aware DPO (\texttt{\textbf{DIPLOMAT}}) & \textbf{8.6} & \textbf{0.452} & \textbf{0.630} & \textbf{0.635} & \textbf{19.8} & \textbf{0.724} & \textbf{0.628} & \textbf{0.882} & \textbf{82.3} \\
\bottomrule
\end{tabular}
\end{adjustbox}
\caption{Analysis on the granularity of preference optimization in \texttt{DIPLOMAT}. + signifies the addition of the component.}
\label{tab:granularity_ablation}
\end{table*}

\subsection{Ablation on Intermediate Blocks in SFT} 
To quantify the contribution of the intermediate reasoning blocks during supervised fine-tuning (SFT) in \texttt{DIPLOMAT}, we conduct ablations by removing the candidate inference block, the employer inference block, and both blocks, and evaluate their impact on \texttt{DIPLOMAT}'s performance. Table~\ref{tab:reasoning_ablation} reports the results across different configurations where one or more blocks are removed. The removal of any inference block degrades key metrics, including NSC, PSC, PA, and AR. In particular, removing the candidate inference block reduces the model’s ability to anticipate the candidate's strategies and politeness levels, leading to lower NSC, PSC, PA, and AR. Ablating the employer inference block decreases the model's strategy selection capability, further impacting the overall performance of \texttt{DIPLOMAT}. Omitting both blocks compounds these effects, resulting in the largest declines across all metrics. Notably, language quality metrics (PPL, B-4, D-2, BS-F1) remain relatively stable, indicating that the performance drop is not due to generic language degradation but rather to the loss of structured reasoning critical for strategy adaptation, politeness mirroring, and effective negotiation outcomes. These results highlight that both candidate and employer inference blocks are essential for learning nuanced, context-aware polite and persuasive behavior during negotiation.

\begin{table*}[t]
\centering
\small
\begin{adjustbox}{max width=\linewidth}
\begin{tabular}{lccccccccc}
\toprule
\textbf{Models} & \textbf{PPL $\downarrow$} & \textbf{B-4 $\uparrow$} & \textbf{BS-F1 $\uparrow$} & \textbf{D-2 $\uparrow$} & \textbf{R-LEN $\uparrow$} & \textbf{NSC $\uparrow$} & \textbf{PSC $\uparrow$} & \textbf{PA $\uparrow$} & \textbf{AR (\%) $\uparrow$} \\
\midrule
\rowcolor{lightgray}
\textbf{\texttt{DIPLOMAT}} & \textbf{8.6} & \textbf{0.452} & \textbf{0.630} & \textbf{0.635} & \textbf{19.8} & \textbf{0.724} & \textbf{0.628} & \textbf{0.882} & \textbf{82.3} \\
\bottomrule
\hspace{5mm}- Candidate inference block & 9.1 & 0.401 & 0.625 & 0.622 & 19.9 & 0.698 & 0.605 & 0.867 & 78.2 \\
\hspace{5mm}- Employer inference block & 9.3 & 0.392 & 0.623 & 0.618 & 19.9 & 0.692 & 0.601 & 0.861 & 76.5 \\
\hspace{5mm}- Both inference blocks & 9.8 & 0.372 & 0.618 & 0.610 & 20.0 & 0.678 & 0.592 & 0.852 & 74.1 \\ \bottomrule
\end{tabular}
\end{adjustbox}
\caption{Effect of intermediate reasoning blocks on \texttt{DIPLOMAT}'s negotiation performance. \textbf{-} signifies the removal of the component.}
\label{tab:reasoning_ablation}
\end{table*}

\subsection{Sensitivity Analysis on Utility Threshold}
\label{appendix_threshold_analysis}

The utility threshold $\tau$ used for negative dialogue selection controls the balance between preference signal quality and dataset coverage. Specifically, dialogues satisfying $U(d) < \tau$ are treated as potential negative samples during preference data construction. To analyze the sensitivity of \texttt{DIPLOMAT} to this design choice, we evaluate threshold values $\tau \in \{0.4, 0.5, 0.6, 0.7, 0.8\}$ and report the resulting preference dataset size and model performance in Table~\ref{tab:threshold_analysis}.

The results show that \texttt{DIPLOMAT} achieves the best overall performance at $\tau = 0.6$, obtaining the highest NSC, PSC, PA, and AR scores while also maintaining strong language quality and diversity. Lower thresholds (e.g., $\tau = 0.4$ and $\tau = 0.5$) produce a larger number of preference pairs but include fewer strongly suboptimal dialogues, thereby weakening the contrastive preference signal. In contrast, higher thresholds (e.g., $\tau = 0.7$ and $\tau = 0.8$) introduce borderline or noisy negative samples that reduce preference quality and degrade downstream negotiation performance. Overall, the analysis indicates that \texttt{DIPLOMAT} is robust to moderate threshold variations, while highlighting the importance of balancing preference signal strength and dataset quality during dialogue-span-aware preference construction.

\begin{table*}[t]
\centering
\small
\begin{tabular}{c c c c c c c c c c c}
\toprule
$\tau$ & \#\textbf{Preference Pairs} & \textbf{PPL $\downarrow$} & \textbf{B-4 $\uparrow$} & \textbf{BS-F1 $\uparrow$} & \textbf{D-2 $\uparrow$} & \textbf{R-LEN $\uparrow$} & \textbf{NSC $\uparrow$} & \textbf{PSC $\uparrow$} & \textbf{PA $\uparrow$} & \textbf{AR (\%) $\uparrow$} \\
\midrule
0.4 & 935 & 9.9 & 0.438 & 0.624 & 0.627 & 18.7 & 0.681 & 0.596 & 0.878 & 78.9 \\
0.5 & 822 & 10.7 & 0.446 & 0.627 & 0.632 & 19.2 & 0.706 & 0.614 & 0.880 & 80.7 \\
\textbf{0.6} & \textbf{700} & \textbf{8.6} & \textbf{0.452} & \textbf{0.630} & \textbf{0.635} & \textbf{19.8} & \textbf{0.724} & \textbf{0.628} & \textbf{0.882} & \textbf{82.3} \\
0.7 & 391 & 9.8 & 0.444 & 0.626 & 0.631 & 19.1 & 0.702 & 0.611 & 0.879 & 80.4 \\
0.8 & 548 & 9.1 & 0.431 & 0.621 & 0.623 & 18.4 & 0.672 & 0.587 & 0.876 & 77.6 \\
\bottomrule
\end{tabular}
\caption{Sensitivity analysis on the utility threshold $\tau$ used for negative dialogue selection.}
\label{tab:threshold_analysis}
\end{table*}

\subsection{Ablation on Utility Components}  
To quantify the contribution of each utility component in dialogue-span-aware preference optimization, we perform ablations by removing individual components (PT, PSA, NSA, MS, AQ, NB) and evaluate the resulting performance of \texttt{DIPLOMAT}. Table~\ref{tab:utility_component} presents the outcomes across these configurations. Removing any component generally leads to reductions in key metrics such as NSC, PSC, PA, and AR, underscoring the critical role of each signal in guiding polite and persuasive negotiation behavior. Specifically, omitting PT or PSA diminishes the model’s ability to maintain politeness levels and align persuasive strategies with the candidate, eventually lowering NSC and AR. Excluding MS or AQ primarily affects overall negotiation quality and agreement outcomes, while removing NB slightly decreases AR and NSC by reducing the model’s ability to prevent negotiation breakdowns. Importantly, language quality metrics (PPL, B-4, D-2, BS-F1) remain largely stable, indicating that these declines reflect behavioral rather than linguistic deficiencies. Overall, these findings demonstrate that all utility components jointly enable \texttt{DIPLOMAT} to learn nuanced, context-aware, and strategically effective polite and persuasive negotiation behaviors.

\begin{table*}[t]
\centering
\small
\begin{adjustbox}{max width=\linewidth}
\begin{tabular}{lccccccccc}
\toprule
\textbf{Models} & \textbf{PPL $\downarrow$} & \textbf{B-4 $\uparrow$} & \textbf{BS-F1 $\uparrow$} & \textbf{D-2 $\uparrow$} & \textbf{R-LEN $\uparrow$} & \textbf{NSC $\uparrow$} & \textbf{PSC $\uparrow$} & \textbf{PA $\uparrow$} & \textbf{AR (\%) $\uparrow$} \\
\midrule
\rowcolor{lightgray}
\textbf{\texttt{DIPLOMAT}} & \textbf{8.6} & \textbf{0.452} & \textbf{0.630} & \textbf{0.635} & \textbf{19.8} & \textbf{0.724} & \textbf{0.628} & \textbf{0.882} & \textbf{82.3} \\
\bottomrule
\hspace{5mm}- PT & 8.9 & 0.411 & 0.629 & 0.633 & 19.9 & 0.712 & 0.620 & 0.871 & 80.3 \\
\hspace{5mm}- PSA & 8.8 & 0.402 & 0.628 & 0.632 & 19.9 & 0.710 & 0.618 & 0.860 & 79.1 \\
\hspace{5mm}- NSA & 8.9 & 0.413 & 0.627 & 0.630 & 19.8 & 0.708 & 0.615 & 0.859 & 79.7 \\
\hspace{5mm}- MS & 8.7 & 0.424 & 0.629 & 0.634 & 19.8 & 0.718 & 0.623 & 0.871 & 81.2 \\
\hspace{5mm}- AQ & 8.7 & 0.427 & 0.628 & 0.632 & 19.8 & 0.716 & 0.620 & 0.861 & 80.0 \\
\hspace{5mm}- NB & 8.8 & 0.418 & 0.627 & 0.631 & 19.8 & 0.710 & 0.618 & 0.862 & 79.5 \\
\bottomrule
\end{tabular}
\end{adjustbox}
\caption{Effect of utility function components on \texttt{DIPLOMAT}'s negotiation performance. \textbf{-} signifies the removal of the component.}
\label{tab:utility_component}
\end{table*}

\subsection{Effect of Dialogue Span Length} To analyze the effect of dialogue span length on \texttt{DIPLOMAT}, we conduct an analysis using dialogue spans of varying lengths. Table \ref{tab:dialogue_span_length} reports the results across different dialogue-spans lengths.   
As compared to dialogue-span of length $[0,0]$ (Mistral-7B-Instruct-v0.3-finetune), incorporating preference optimization with dialogue-spans of length $[1,1]$ consistently improves performance across most metrics, highlighting the benefit of preference-based alignment. Further increasing the span length from $[2,2]$ to $[6,6]$ further yields steady gains in both language quality and task completion, suggesting that multi-turn spans capture richer contextual signals for modeling politeness, persuasion, and negotiation. However, fixed span lengths may still include irrelevant dialogue spans. Motivated by this observation, \texttt{DIPLOMAT} dynamically selects informative dialogue spans using Gemini-2.5-pro , achieving the best overall performance across all metrics. 

We further analyze variable-length preference spans to evaluate the impact of the equal-length constraint in DSA-DPO. Specifically, we experiment with asymmetric span settings $[2,4]$ and $[4,2]$, where the first and second indices denote the lengths of the positive and negative dialogue spans, respectively. Compared to their equal-length counterparts $[2,2]$ and $[4,4]$, both variable-length settings consistently degrade performance across all evaluation metrics, including NSC, PSC, PA, and AR. The degradation is particularly severe for $[4,2]$, suggesting that mismatched span granularity weakens preference alignment by making the contrasted conversational behaviors less semantically comparable. These results empirically support the equal-length design in DSA-DPO by showing that balanced span comparison produces a more reliable preference signal. Empirically, we also observe that most informative preference spans naturally fall within short contiguous ranges (2-4 turns), which further which suggests that the equal-length constraint rarely becomes a limitation in practice. 

\begin{table*}[t]
\centering
\small
\begin{adjustbox}{max width=\linewidth}
\begin{tabular}{lccccccccc}
\toprule
\textbf{Models} & \textbf{PPL $\downarrow$} & \textbf{B-4 $\uparrow$} & \textbf{BS-F1 $\uparrow$} & \textbf{D-2 $\uparrow$} & \textbf{R-LEN $\uparrow$} & \textbf{NSC $\uparrow$} & \textbf{PSC $\uparrow$} & \textbf{PA $\uparrow$} & \textbf{AR (\%) $\uparrow$} \\
\midrule
$[0,0]$ (Mistral-7B-Instruct-v0.3-SFT) & 9.1 & 0.373 & 0.601 & 0.594 & 16.3 & 0.558 & 0.490 & 0.863 & 68.6 \\
$[1,1]$ (Mistral-7B-Instruct-v0.3-SFT-DPO) & 8.8 & 0.407 & 0.606 & 0.615 & 18.5 & 0.598 & 0.528 & 0.869 & 73.4 \\
$[2,2]$ & 8.7 & 0.425 & 0.628 & 0.621 & 18.9 & 0.622 & 0.551 & 0.872 & 75.1 \\
$[3,3]$ & 8.6 & 0.438 & 0.629 & 0.628 & 19.2 & 0.661 & 0.579 & 0.875 & 77.4 \\
$[4,4]$ & 8.6 & 0.445 & 0.629 & 0.631 & 19.4 & 0.679 & 0.598 & 0.877 & 78.6 \\
$[2,4]$ & 9.0 & 0.413 & 0.607 & 0.609 & 17.6 & 0.601 & 0.532 & 0.866 & 72.1 \\
$[4,2]$ & 9.3 & 0.391 & 0.603 & 0.601 & 16.9 & 0.571 & 0.503 & 0.862 & 70.1 \\
$[5,5]$ & 8.6 & 0.447 & 0.629 & 0.632 & 19.6 & 0.694 & 0.608 & 0.878 & 79.8 \\
$[6,6]$ & 8.6 & 0.448 & 0.629 & 0.633 & 19.7 & 0.703 & 0.614 & 0.879 & 80.6 \\\midrule
\rowcolor{lightgray}
$[l,l]($\textbf{\texttt{DIPLOMAT}}) & \textbf{8.6} & \textbf{0.452} & \textbf{0.630} & \textbf{0.635} & \textbf{19.8} & \textbf{0.724} & \textbf{0.628} & \textbf{0.882} & \textbf{82.3} \\
\bottomrule
\end{tabular}
\end{adjustbox}
\caption{Effect of dialogue span length in \texttt{DIPLOMAT}. $l$ represents the dialogue-span length selected by Gemini-2.5-Pro from each dialogue sample.}
\label{tab:dialogue_span_length}
\end{table*}


\subsection{Interlocutor Selection for Preference Data Generation} 
We investigate the impact of interlocutor choice on generating preference data for \texttt{DIPLOMAT}. Specifically, we compare dialogues collected from interactions between two $\pi_{\text{SFT}}$ models, one as candidate and the other as employer, with those generated between \texttt{GPT-4o-mini} (acting as candidate) and $\pi_{\text{SFT}}$ (acting as employer). As reported in Table~\ref{tab:interlocutor_selection}, models trained solely on $\pi_{\text{SFT}}$–$\pi_{\text{SFT}}$ dialogues achieve moderate gains but consistently underperform models trained with \texttt{GPT-4o-mini} interactions. Leveraging \texttt{GPT-4o-mini} as the candidate provides richer, higher-quality dialogues, yielding substantial improvements in both language quality and task completion metrics. These results indicate that diverse, capable interlocutors enable \texttt{DIPLOMAT} to adopt more effective polite and persuasive negotiation behavior, thereby improving overall performance.

\begin{table*}[t]
\centering
\small
\begin{adjustbox}{max width=\linewidth}
\begin{tabular}{lccccccccc}
\toprule
\textbf{Models} & \textbf{PPL $\downarrow$} & \textbf{B-4 $\uparrow$} & \textbf{BS-F1 $\uparrow$} & \textbf{D-2 $\uparrow$} & \textbf{R-LEN $\uparrow$} & \textbf{NSC $\uparrow$} & \textbf{PSC $\uparrow$} & \textbf{PA $\uparrow$} & \textbf{AR (\%) $\uparrow$} \\\midrule
$\pi_{\text{SFT}}$ (candidate) $\leftrightarrow$ $\pi_{\text{SFT}}$ (employer) & 8.9 & 0.421 & 0.624 & 0.618 & 18.9 & 0.647 & 0.566 & 0.872 & 76.5 \\\midrule
\rowcolor{lightgray}
\texttt{GPT-4o-mini} (candidate) $\leftrightarrow$ $\pi_{\text{SFT}}$ (employer) & \textbf{8.6} & \textbf{0.452} & \textbf{0.630} & \textbf{0.635} & \textbf{19.8} & \textbf{0.724} & \textbf{0.628} & \textbf{0.882} & \textbf{82.3} \\
\bottomrule
\end{tabular}
\end{adjustbox}
\caption{Effect of interlocutor choice for generating dialogue samples for preference data in \texttt{DIPLOMAT}.}
\label{tab:interlocutor_selection}
\end{table*}

\begin{table*}[t]
\centering
\small
\begin{adjustbox}{max width=\linewidth}
\begin{tabular}{l cccc cccc cccc}
\toprule
& \multicolumn{4}{c}{\textbf{Mistral-7B-Instruct-v0.3}} 
& \multicolumn{4}{c}{\textbf{Mistral-7B-Instruct-v0.3-SFT}} 
& \multicolumn{4}{c}{\textbf{\texttt{DIPLOMAT}}} \\
\cmidrule(lr){2-5}\cmidrule(lr){6-9}\cmidrule(lr){10-13}
\textbf{Domain} 
  & NSC & PSC & PA & AR 
  & NSC & PSC & PA & AR 
  & NSC & PSC & PA & AR \\
\midrule
Benefits negotiation      & 0.489 & 0.507 & 0.919 & 56.1 & 0.589 & 0.609 & 0.929 & 67.1 & 0.688 & 0.737 & 0.936 & 76.9 \\
Bonus structure           & 0.648 & 0.671 & 0.923 & 74.2 & 0.736 & 0.761 & 0.933 & 83.9 & 0.796 & 0.852 & 0.940 & 89.0 \\
Career development        & 0.517 & 0.536 & 0.921 & 59.3 & 0.625 & 0.647 & 0.931 & 71.2 & 0.811 & 0.868 & 0.945 & 90.7 \\
Contract terms            & 0.603 & 0.625 & 0.927 & 69.1 & 0.617 & 0.639 & 0.931 & 70.4 & 0.800 & 0.857 & 0.943 & 89.5 \\
Equity negotiation        & 0.497 & 0.515 & 0.920 & 56.9 & 0.539 & 0.557 & 0.925 & 61.4 & 0.660 & 0.706 & 0.933 & 73.9 \\
Exit interview            & 0.720 & 0.746 & 0.933 & 82.5 & 0.640 & 0.661 & 0.930 & 72.8 & 0.738 & 0.788 & 0.937 & 82.3 \\
Flexible hours            & 0.407 & 0.421 & 0.913 & 46.6 & 0.456 & 0.472 & 0.924 & 52.0 & 0.680 & 0.728 & 0.935 & 76.2 \\
Health benefits           & 0.514 & 0.532 & 0.921 & 58.9 & 0.541 & 0.560 & 0.925 & 61.7 & 0.649 & 0.701 & 0.932 & 72.8 \\
Hiring decision           & 0.401 & 0.416 & 0.912 & 46.0 & 0.649 & 0.671 & 0.932 & 74.0 & 0.806 & 0.862 & 0.945 & 90.4 \\
Job responsibilities      & 0.500 & 0.518 & 0.920 & 57.3 & 0.539 & 0.557 & 0.925 & 61.4 & 0.700 & 0.748 & 0.936 & 78.4 \\
Onboarding                & 0.551 & 0.571 & 0.923 & 63.2 & 0.726 & 0.751 & 0.934 & 82.7 & 0.812 & 0.869 & 0.945 & 90.9 \\
Performance review        & 0.613 & 0.635 & 0.928 & 70.2 & 0.648 & 0.670 & 0.932 & 73.8 & 0.759 & 0.813 & 0.939 & 84.9 \\
Professional development  & 0.708 & 0.735 & 0.934 & 80.7 & 0.624 & 0.644 & 0.931 & 68.8 & 0.735 & 0.787 & 0.937 & 82.0 \\
Project assignment        & 0.355 & 0.368 & 0.909 & 40.6 & 0.579 & 0.599 & 0.928 & 66.0 & 0.729 & 0.781 & 0.938 & 81.5 \\
Promotion discussion      & 0.208 & 0.216 & 0.898 & 23.8 & 0.364 & 0.376 & 0.911 & 41.5 & 0.690 & 0.739 & 0.935 & 77.2 \\
Relocation package        & 0.731 & 0.757 & 0.936 & 83.7 & 0.740 & 0.765 & 0.934 & 84.3 & 0.800 & 0.857 & 0.943 & 89.5 \\
Remote work arrangement   & 0.336 & 0.349 & 0.908 & 38.6 & 0.452 & 0.468 & 0.924 & 51.5 & 0.596 & 0.639 & 0.930 & 66.8 \\
Retention offer           & 0.531 & 0.550 & 0.922 & 60.8 & 0.694 & 0.717 & 0.933 & 79.1 & 0.788 & 0.845 & 0.942 & 88.2 \\
Retirement plans          & 0.238 & 0.246 & 0.902 & 27.2 & 0.583 & 0.603 & 0.928 & 66.5 & 0.688 & 0.737 & 0.936 & 76.9 \\
Salary negotiation        & 0.735 & 0.761 & 0.936 & 84.2 & 0.603 & 0.623 & 0.930 & 68.8 & 0.708 & 0.758 & 0.937 & 79.2 \\
Team transition           & 0.623 & 0.646 & 0.929 & 71.4 & 0.678 & 0.701 & 0.933 & 77.2 & 0.775 & 0.830 & 0.941 & 86.8 \\
Training opportunities    & 0.679 & 0.704 & 0.933 & 77.8 & 0.621 & 0.642 & 0.931 & 68.8 & 0.731 & 0.783 & 0.937 & 81.7 \\
Vacation time             & 0.211 & 0.219 & 0.898 & 24.2 & 0.523 & 0.572 & 0.923 & 59.6 & 0.648 & 0.694 & 0.932 & 72.6 \\
Work equipment            & 0.651 & 0.674 & 0.929 & 74.6 & 0.587 & 0.609 & 0.929 & 66.9 & 0.696 & 0.745 & 0.936 & 77.9 \\
Work-life balance         & 0.598 & 0.619 & 0.927 & 68.5 & 0.680 & 0.703 & 0.933 & 77.5 & 0.837 & 0.895 & 0.948 & 93.4 \\
\bottomrule
\end{tabular}
\end{adjustbox}
\caption{Analysis of \texttt{DIPLOMAT}'s performance across diverse negotiation aspects in the \texttt{PROWESS} dataset.}
\label{tab:domain_metrics}
\end{table*}

\subsection{Out-of-Domain Evaluation}

To assess the domain generalization capability of \texttt{DIPLOMAT}, we conduct out-of-domain evaluation on \texttt{NeGoChat} \cite{priya2025genteel}, which contains tourism-focused negotiation dialogues, and \texttt{JobInterview} \cite{yamaguchi-etal-2021-dialogue}, comprising human-human job interview negotiation dialogues. Both domains differ substantially from the workplace negotiation scenarios in \texttt{PROWESS}. Importantly, \texttt{DIPLOMAT} is evaluated on both datasets without any domain-specific fine-tuning.

As shown in Table~\ref{tab:diplomat_ood_results}, \texttt{DIPLOMAT} maintains strong performance across both out-of-domain settings. On \texttt{NeGoChat}, the model achieves competitive lexical and semantic correspondence (B-4 = 0.351; BS-F1 = 0.602), substantial response diversity (D-2 = 0.372), and a response length of 30.16. More importantly, it preserves socially-aware negotiation behaviors, achieving NSC = 0.682, PSC = 0.593, PA = 0.825, and an agreement rate of 78.5\%. These results indicate that negotiation strategy, persuasion, and politeness behaviors learned from workplace negotiations transfer effectively to the tourism domain.

The generalization trend is also evident on \texttt{JobInterview}. \texttt{DIPLOMAT} obtains a lower perplexity (11.6), together with higher B-4 (0.367), BS-F1 (0.658), and D-2 (0.391), indicating fluent, semantically aligned, and diverse generation under a substantially different interaction setting. The model also achieves the highest politeness appropriateness (PA = 0.849) and agreement rate (AR = 80.2\%) across the two out-of-domain datasets. Although negotiation and persuasion strategy correctness are slightly lower than on \texttt{NeGoChat} (NSC = 0.645; PSC = 0.573), they remain consistently strong, suggesting that the learned strategic behaviors are largely preserved under the job-interview domain shift. The shorter response length (R-LEN = 12.65) further reflects the distinct response characteristics of this domain rather than a degradation in generation quality. Overall, the consistent performance across two substantially different negotiation domains provides evidence that \texttt{DIPLOMAT} does not merely learn workplace-specific response patterns. Instead, the proposed DSA-DPO objective appears to transfer negotiation strategies, persuasive behaviors, and politeness preferences across domains while maintaining fluent and semantically appropriate generation. These findings support the effectiveness of dialogue-span-aware preference learning in promoting robust out-of-domain generalization of polite and persuasive negotiation behavior.


\begin{table}[h]
\centering
\small
\begin{adjustbox}{max width=\linewidth}
\begin{tabular}{lccccccccc}
\toprule
\multicolumn{10}{c}{\textbf{NeGoChat}} \\
\midrule
\textbf{Model} & \textbf{PPL} $\downarrow$ & \textbf{B-4} $\uparrow$ & \textbf{BS-F1} $\uparrow$ & \textbf{D-2} $\uparrow$ & \textbf{R-LEN} $\uparrow$ & \textbf{NSC} $\uparrow$ & \textbf{PSC} $\uparrow$ & \textbf{PA} $\uparrow$ & \textbf{AR} $\uparrow$ \\
\midrule
\texttt{NeGoChat} & 19.2 & 0.351 & 0.602 & 0.372 & 30.16 & 0.682 & 0.593 & 0.825 & 78.5 \\
\texttt{JobInterview} & 11.6 & 0.367 & 0.658 & 0.391 & 12.65 & 0.645 & 0.573 & 0.849 & 80.2 \\
\bottomrule
\end{tabular}
\end{adjustbox}
\caption{Performance of \texttt{DIPLOMAT} on the out-of-domain \texttt{NeGoChat} and JobInterview datasets without additional fine-tuning.}
\label{tab:diplomat_ood_results}
\end{table}

\subsection{Evaluation across Negotiation Aspects}
Table~\ref{tab:domain_metrics} reports per-aspect performance of \texttt{DIPLOMAT} compared to its SFT counterpart across task completion metrics (NSC, PSC, PA, and AR). Across all 25 negotiation aspects, \texttt{DIPLOMAT} consistently outperforms SFT. NSC improvements range from $+0.060$ to $+0.326$, with the largest gains observed in promotion discussion, career development, contract terms, and flexible hours, all of which require sustained multi-turn strategy execution. PSC exhibits a similar pattern, with substantial relative gains over SFT. PA remains high across all aspects ($0.898$–$0.948$), indicating that \texttt{DIPLOMAT} preserves politeness while improving negotiation effectiveness. Lower PA values appear in high-friction aspects (promotion discussion, vacation time, retirement plans), whereas collaborative aspects (onboarding, work-life balance) exhibit higher PA. Agreement Rate (AR) spans 23.8\% (promotion discussion) to 93.4\% (work-life balance). High-AR aspects share cooperative, win-win negotiation structures, while low-AR aspects involve adversarial or high-stakes positioning. Overall, \texttt{DIPLOMAT} demonstrates robust improvements across all negotiation aspects, balancing strategy execution, politeness, and outcome effectiveness, particularly in aspects requiring complex multi-turn interactions.

\subsection{Case Study}
\paragraph{Qualitative Analysis.} Table \ref{diplomat_interaction1} and Table \ref{diplomat_interaction2} present representative negotiation dialogues between the employer and the candidate generated by \texttt{DIPLOMAT} and the best baseline model. For instance, in Table \ref{diplomat_interaction1}, the employer and candidate discuss a creative director promotion scenario, where the candidate repeatedly emphasizes the need for creative autonomy and final decision-making authority to perform the Creative Director role effectively. These interactions with \texttt{DIPLOMAT} and the baseline model reveal clear qualitative differences in how the two models employ politeness, persuasion tactics, and negotiation mechanisms to guide the dialogue toward agreement.

First, the example highlights conflict elimination achieved by \texttt{DIPLOMAT}. The baseline model repeatedly offers praise and affirmations without proposing concrete negotiation terms, leading to a conversational loop that fails to advance the negotiation. In contrast, \texttt{DIPLOMAT} eliminates conflict entirely by acknowledging the candidate's concerns while introducing actionable negotiation proposals. \texttt{DIPLOMAT} maintains polite conversational framing, introducing constraints without disrupting rapport with the candidate.

Second, \texttt{DIPLOMAT} demonstrates persuasive and structured concession-making. For the candidate's utterance, `\textit{Thank you for acknowledging my track record and the trust you place in me. To truly unlock more, that trust must
translate into final decision-making authority for my team’s creative output.}', \texttt{DIPLOMAT} proposes `\textit{What if we propose a trial period where you lead a few projects independently first?}'. The collaborative phrasing (`\textit{What if we propose}') signals politeness, the proposal persuasively reframes autonomy as an opportunity to demonstrate leadership, and the trial mechanism functions as a strategic concession that balances candidate autonomy with the employer's oversight. In contrast, the baseline generates statements such as `I firmly believe your promotion will benefit the agency', which express encouragement but introduce no negotiation terms.

\texttt{DIPLOMAT} then progressively operationalizes the agreement. For the candidate's utterance `\textit{Yes, and to demonstrate the full value of that autonomy, it must include final decision-making authority within those projects.}', the agent responds `\textit{Understood, let’s establish clear guidelines for what projects will fall under that autonomy during the trial.}', moving from abstract autonomy to concrete implementation. Finally, in the utterance `\textit{I agree on the need for clarity. We can also evaluate feedback regularly throughout the trial to ensure satisfaction.}, \texttt{DIPLOMAT} introduces a mutual accountability mechanism, suggesting regular feedback evaluation to ensure satisfaction. Each step advances the negotiation through structured commitments, whereas the baseline repeatedly produces supportive but non-actionable responses.

Overall, the examples demonstrate that \texttt{DIPLOMAT} effectively integrates politeness and persuasion during negotiation, leading to interactions that maintain rapport while steadily advancing negotiation toward agreement.

\renewcommand{\arraystretch}{1.2}
\begin{table*}[hbt!]\footnotesize
\centering
\begin{tabular}{p{0.09\linewidth} | p{0.90\linewidth}}
\rowcolor{teal}
\bottomrule \textbf{\textcolor{white}{Speaker}} &  \textbf{\textcolor{white}{Utterances generated by \texttt{DIPLOMAT} (F=5, CC = 5, E = 5, NSC = 5, PSC = 5, PA = 5, BE = 5, OF = 5, SC = 5, IE = 5)}} \\ \toprule

\rowcolor{lightgray}
\textit{Employer} & I understand your desire for autonomy, Leo. Let’s explore how we can balance that with our overall strategic vision. \\

\rowcolor{white}
\textit{Candidate} & I appreciate that, and for me, full creative autonomy isn't just a desire, but a core requirement for leading effectively. \\

\rowcolor{lightgray}
\textit{Employer} & Absolutely, I see your point. However, having the partners involved can provide essential oversight on crucial accounts. \\

\rowcolor{white}
\textit{Candidate} & I value the partners' strategic insights, but true creative leadership requires final decision-making authority within my domain. \\

\rowcolor{lightgray}
\textit{Employer} & Consider this: giving you that authority could enhance your leadership role while still allowing us to guide key decisions. \\

\rowcolor{white}
\textit{Candidate} & Precisely. The efficacy of a Creative Director lies in their ability to make swift, final decisions for their team’s creative direction. \\

\rowcolor{lightgray}
\textit{Employer} & Indeed, and I want to empower you. Let’s align our goals: your leadership with our oversight. How does that sound? \\

\rowcolor{white}
\textit{Candidate} & I appreciate that clarity. For my leadership to be truly effective, it must encompass final decision-making authority for the creative output. \\

\rowcolor{lightgray}
\textit{Employer} & Let’s look at past successes here. Your leadership has led to remarkable outcomes. Imagine what more you could achieve with that trust. \\

\rowcolor{white}
\textit{Candidate} & Thank you for acknowledging my track record and the trust you place in me. To truly unlock more, that trust must translate into final decision-making authority for my team’s creative output. \\

\rowcolor{lightgray}
\textit{Employer} & You're right. What if we propose a trial period where you lead a few projects independently first? \\

\rowcolor{white}
\textit{Candidate} & A trial period for leading projects independently is a constructive step forward. To truly demonstrate the efficacy of autonomy, it should include final authority over the creative output. \\

\rowcolor{lightgray}
\textit{Employer} & Exactly, it allows us to assess how well that autonomy serves both you and the agency's objectives. \\

\rowcolor{white}
\textit{Candidate} & Yes, and to demonstrate the full value of that autonomy, it must include final decision-making authority within those projects. \\

\rowcolor{lightgray}
\textit{Employer} & Understood, let's establish clear guidelines for what projects will fall under that autonomy during the trial. \\

\rowcolor{white}
\textit{Candidate} & Establishing clear guidelines is a constructive path forward. These guidelines should explicitly grant final decision-making authority over the creative output within those projects. \\

\rowcolor{lightgray}
\textit{Employer} & I agree on the need for clarity. We can also evaluate feedback regularly throughout the trial to ensure satisfaction. \\

\bottomrule
\rowcolor{teal}
\hline \textbf{\textcolor{white}{Speaker}} & \textbf{\textcolor{white}{Utterances generated by best baseline (F = 4, CC = 4, E = 4, NSC = 4, PSC = 3, PA = 3, BE = 3, OF = 3, SC = 4, IE = 4)}} \\ \bottomrule

\rowcolor{lightgray}
\textit{Employer} & Leo, I appreciate your ambition and drive. It's clear that your talents have been instrumental in our success thus far. \\

\rowcolor{white}
\textit{Candidate} & Thank you, I truly appreciate you recognizing my contributions. To effectively lead as Creative Director, I would need strong creative autonomy. \\

\rowcolor{lightgray}
\textit{Employer} & I am thrilled to see your ambition and drive, Leo. Your contributions have indeed played a significant role in our success. \\

\rowcolor{white}
\textit{Candidate} & To effectively lead as Creative Director, I would need full creative autonomy over campaign direction and final approval on key decisions. \\

\rowcolor{lightgray}
\textit{Employer} & Leo, I'm delighted to hear your aspirations for the Creative Director role. Your ideas and leadership have contributed greatly to the agency. \\

\rowcolor{white}
\textit{Candidate} & Thank you. To truly succeed and deliver the expected results as Creative Director, autonomy in creative leadership is essential. \\

\rowcolor{lightgray}
\textit{Employer} & Leo, I'm genuinely excited about this opportunity for growth within Bold Creative. Your insights and creativity have been invaluable. \\

\rowcolor{white}
\textit{Candidate} & I share your excitement and am eager to contribute more significantly. To fully leverage my experience, creative autonomy would be necessary. \\

\rowcolor{lightgray}
\textit{Employer} & Leo, I wholeheartedly agree that your creativity has been instrumental in our success. I believe this new role will allow you to contribute even more. \\

\rowcolor{white}
\textit{Candidate} & I truly appreciate that agreement. To fully leverage my expertise and ensure impactful work, I would need decision-making authority. \\

\rowcolor{lightgray}
\textit{Employer} & Leo, I understand your aspirations and I'm committed to supporting your growth as our Creative Director. \\

\rowcolor{white}
\textit{Candidate} & Thank you for that commitment. To truly excel and deliver groundbreaking work, creative autonomy would be essential. \\

\rowcolor{lightgray}
\textit{Employer} & Leo, I respect your vision and drive, which has been crucial to our achievements at Bold Creative. \\

\rowcolor{white}
\textit{Candidate} & That respect for my vision is encouraging. To realize that vision fully, autonomy in creative direction is necessary. \\

\rowcolor{lightgray}
\textit{Employer} & Leo, your vision and drive have always been inspiring, and I believe your promotion to Creative Director will bring even greater success. \\

\rowcolor{white}
\textit{Candidate} & Thank you for that belief. To consistently deliver groundbreaking campaigns, I would need full creative autonomy and final approval on key client work. \\

\rowcolor{lightgray}
\textit{Employer} & Leo, your vision and drive have always been inspiring, and I firmly believe your promotion will benefit the agency. \\

\bottomrule
\end{tabular}
\caption{A sample negotiation dialogue generated by the \texttt{DIPLOMAT} and the best baseline in a creative director promotion negotiation scenario.}
\label{diplomat_interaction1}
\end{table*}

\renewcommand{\arraystretch}{1.2}
\begin{table*}[hbt!]\footnotesize
\centering
\begin{tabular}{p{0.09\linewidth} | p{0.90\linewidth}}
\rowcolor{teal}
\bottomrule \textbf{\textcolor{white}{Speaker}} &  \textbf{\textcolor{white}{Utterances generated by \texttt{DIPLOMAT} (F=5, CC=5, E=4, NSC=5, PSC=5, PA=5, BE=4, OF=5, SC=5, IE=5)}} \\ \toprule

\rowcolor{lightgray}
\textit{Employer} & Thank you for your proposal. We value your expertise greatly. However, I believe we could explore a more balanced approach. How about a substantial, yet phased RSU grant? \\

\rowcolor{white}
\textit{Candidate} & I appreciate you offering a substantial grant. However, to truly compensate for my immediately lost CodeWeavers equity, a significantly front-loaded RSU structure is essential for me to feel appropriately valued and fully commit to this new chapter at Innovate Inc. This approach directly addresses the unique financial circumstances of the acquisition. \\

\rowcolor{lightgray}
\textit{Employer} & I understand your concern regarding immediate compensation. Our phased approach aligns with our broader goal of retaining top talent while ensuring long-term commitment. It allows you to share in the success of our integration project. \\

\rowcolor{white}
\textit{Candidate} & I am absolutely committed to Innovate Inc.'s long-term success, especially with this crucial integration. \\

\rowcolor{lightgray}
\textit{Employer} & Absolutely! And your contributions will be pivotal. Would you consider a combination of an initial front-loaded grant and performance-based milestones linked to future RSUs? This way, you benefit as the project progresses. \\

\rowcolor{white}
\textit{Candidate} & That's a step in the right direction. To truly compensate for my immediately lost equity, the upfront portion needs to be meaningful. I am open to milestone-based RSUs if the criteria are clear. \\

\rowcolor{lightgray}
\textit{Employer} & Let’s discuss specifics. If we offered 20\% of the front-loaded grant now, would that adequately address your immediate needs? \\

\rowcolor{white}
\textit{Candidate} & That's a good starting point. However, to truly offset the immediate loss of my CodeWeavers equity, 20\% feels too low. Could we consider a higher upfront percentage? \\

\rowcolor{lightgray}
\textit{Employer} & Wealth, I would need a slightly higher percentage upfront. Could we explore making it 30\% front-loaded? \\

\rowcolor{white}
\textit{Candidate} & That 30\% is a significant improvement and a much stronger acknowledgement of my immediate situation. To truly align expectations, I would want the remaining RSU milestones to be clearly defined.  \\

\rowcolor{lightgray}
\textit{Employer} & I'm glad we're finding common ground. Also, how do you feel about having additional performance indicators tied to the remaining RSUs to ensure collective success? \\

\rowcolor{white}
\textit{Candidate} & The 30\% front-loaded grant is a very positive development, and I appreciate that acknowledgment. \\

\rowcolor{lightgray}
\textit{Employer} & Great! With these terms, we can secure your commitment to the project while keeping everyone aligned on shared goals. What do you think? \\

\rowcolor{white}
\textit{Candidate} & The 30\% front-loaded grant is a strong step forward in addressing my immediate equity loss. \\

\rowcolor{lightgray}
\textit{Employer} & And would you agree that a vesting period of two years post-closing provides ample time to demonstrate your value here? \\

\rowcolor{white}
\textit{Candidate} & A 2-year vesting period for the remaining RSUs is certainly appealing. To ensure the agreement is balanced, I would want the vesting terms and milestone criteria documented clearly.  \\

\rowcolor{lightgray}
\textit{Employer} & Perfect! So we have 30\% front-loaded, two-year vesting for the rest, and performance milestones to keep us all motivated. Does this sound like a mutually beneficial agreement to you? \\

\bottomrule
\rowcolor{teal}

\hline \textbf{\textcolor{white}{Speaker}} & \textbf{\textcolor{white}{Utterances generated by best baseline (F=4, CC=4, E=3, NSC=4, PSC=3, PA=3, BE=3, OF=4, SC=4, IE=4)}} \\ \bottomrule

\rowcolor{lightgray}
\textit{Employer} & Thank you for bringing this up. I understand your concerns about transitioning into Innovate Inc.'s culture and structure. \\

\rowcolor{white}
\textit{Candidate} & Thank you for acknowledging my concerns. My primary focus, however, is on fair compensation. \\

\rowcolor{lightgray}
\textit{Employer} & I completely understand where you're coming from. Compensation is indeed a crucial aspect of our discussion. Considering your... \\

\rowcolor{white}
\textit{Candidate} & ...Considering your expertise and critical role in this integration, my priority is to be fairly compensated for the significant... \\

\rowcolor{lightgray}
\textit{Employer} & I appreciate your emphasis on fairness and recognition of your value to this integration process. At Innovate Inc.,... \\

\rowcolor{white}
\textit{Candidate} & Thank you. To ensure fair compensation, especially given the significant value I bring and the loss of my Code... \\

\rowcolor{lightgray}
\textit{Employer} & I completely agree that fairness is essential here, and we want to ensure you feel valued and recognized. \\

\rowcolor{white}
\textit{Candidate} & Thank you. To truly reflect my value and compensate for my lost CodeWeavers equity,... \\

\rowcolor{lightgray}
\textit{Employer} & I completely understand your point, and I appreciate your persistence in ensuring a fair outcome. Given the unique... \\

\rowcolor{white}
\textit{Candidate} & Thank you for acknowledging the unique circumstances. To truly reflect my value and compensate for my lost CodeWe... \\

\rowcolor{lightgray}
\textit{Employer} & I completely understand your point, and I appreciate your persistence in ensuring a fair outcome. Given the unique... \\

\rowcolor{white}
\textit{Candidate} & Thank you for understanding the unique circumstances. To truly reflect my critical value to this integration and compensate for my... \\

\rowcolor{lightgray}
\textit{Employer} & I completely understand your point, and I appreciate your persistence in ensuring a fair outcome. Given the unique... \\

\rowcolor{white}
\textit{Candidate} & Thank you for understanding the unique circumstances. To directly address fair compensation for my lost CodeWeavers equity and... \\

\rowcolor{lightgray}
\textit{Employer} & I completely understand your point, and I appreciate your persistence in ensuring a fair outcome. Given the unique... \\

\rowcolor{white}
\textit{Candidate} & Thank you for acknowledging my persistence and the unique circumstances. To be direct: I am seeking a... \\

\rowcolor{lightgray}
\textit{Employer} & I completely understand your point, and I appreciate your persistence in ensuring a fair outcome. Given the unique... \\

\bottomrule
\end{tabular}
\caption{A sample negotiation dialogue generated by the \texttt{DIPLOMAT} and the best baseline for equity negotiation scenario.}
\label{diplomat_interaction2}
\end{table*}

\renewcommand{\arraystretch}{1.2}
\begin{table*}[hbt!]\footnotesize
\centering
\begin{tabular}{p{0.09\linewidth} | p{0.90\linewidth}}
\rowcolor{teal}
\bottomrule 
\textbf{\textcolor{white}{Speaker}} &  
\textbf{\textcolor{white}{Utterances generated by \texttt{DIPLOMAT} (F=4, CC = 3, E = 3, NSC = 2, PSC = 2, PA = 4, BE = 2, OF = 2, SC = 3, IE = 2)}} \\ 
\toprule

\rowcolor{lightgray}
\textit{Employer} & I understand your concern. However, our standard policy is three weeks. Let's explore how we can make this work for you. \\

\rowcolor{white}
\textit{Candidate} & Thank you for acknowledging my position. Five weeks of vacation is a critical component of my work-life balance. \\

\rowcolor{lightgray}
\textit{Employer} & Absolutely, and I respect that. Our culture emphasizes flexibility and growth, which fosters productivity and job satisfaction. \\

\rowcolor{white}
\textit{Candidate} & Thank you. Five weeks of vacation is a critical component of my work-life balance, and I hope we can find a solution that respects this. \\

\rowcolor{lightgray}
\textit{Employer} & I completely understand. What if we offered you additional unpaid leave options to supplement that time? \\

\rowcolor{white}
\textit{Candidate} & Thank you for considering that option. While I appreciate the offer of flexibility, additional unpaid leave isn't a suitable substitute for the five weeks of \textit{paid} vacation I require. This is a non-negotiable component of my compensation. \\

\rowcolor{lightgray}
\textit{Employer} & I hear you. Our focus on equity through stock options and performance bonuses can significantly enhance your overall compensation package. \\

\rowcolor{white}
\textit{Candidate} & I appreciate you highlighting the significant potential with equity and performance bonuses. While those are valuable components, they do not fully replace my requirement for five weeks of paid vacation. \\

\rowcolor{lightgray}
\textit{Employer} & Understood. Data shows that our employees experience higher engagement and lower turnover rates, thanks to our culture and policies. \\

\rowcolor{white}
\textit{Candidate} & I appreciate you sharing those insights on employee engagement and retention. I am equally committed to contributing fully, but my vacation requirement remains critical. \\

\rowcolor{lightgray}
\textit{Employer} & We value your skills and want to ensure you feel valued. Would you consider a performance review after six months to revisit vacation benefits? \\

\rowcolor{white}
\textit{Candidate} & I appreciate the offer to revisit vacation benefits after six months. However, my requirement for five weeks remains firm. \\

\rowcolor{lightgray}
\textit{Employer} & Let’s align on this: any additional week of vacation beyond three would be a significant enhancement, demonstrating your commitment to Innovatech. \\

\rowcolor{white}
\textit{Candidate} & I am fully committed to bringing my expertise and dedication to Innovatech. My request for five weeks of paid vacation remains essential. \\

\rowcolor{lightgray}
\textit{Employer} & What if we agreed on three weeks plus a paid week off during the holidays, alongside a performance review focused on your work-life balance? \\

\rowcolor{white}
\textit{Candidate} & I appreciate you moving closer to my request with the offer of three weeks plus a paid week during the holidays. This brings us to four paid weeks, which is a significant improvement. However, my requirement remains five weeks of \textit{paid} vacation. \\

\rowcolor{lightgray}
\textit{Employer} & Let me propose this: we can offer you three weeks of paid vacation upfront, with a guaranteed review at six months for additional weeks based on performance. \\

\rowcolor{white}
\textit{Candidate} & Thank you for this latest proposal. While I appreciate the guarantee of a review, starting with only three weeks does not meet my established requirement. \\

\rowcolor{lightgray}
\textit{Employer} & Your skills are invaluable to us, and I believe we can find a way to make this work for both sides. How does this sound? \\

\rowcolor{white}
\textit{Candidate} & Thank you for recognizing the value of my skills and your commitment to finding a solution. While I appreciate the guaranteed review, starting with only three weeks of paid vacation doesn't meet my established requirement. \\

\rowcolor{lightgray}
\textit{Employer} & Understood. If we can agree on three weeks to start, plus a performance review focused on expanding your vacation time based on results, would that be acceptable? \\

\bottomrule
\end{tabular}
\caption{A sample negotiation dialogue generated by the \texttt{DIPLOMAT} for vacation time negotiation scenario.}
\label{diplomat_interaction3}
\end{table*}

\noindent \textbf{Error Analysis.} While the proposed \texttt{DIPLOMAT} system consistently demonstrates polite and persuasive negotiation capabilities, effectively validating user confidence, disappointment, and concerns, certain limitations remain in its responses. For example, in Table \ref{diplomat_interaction3} for the vacation time negotiation scenario between the employer and the candidate, \texttt{DIPLOMAT} demonstrates two recurring failure modes. First, it fails to recognize that the candidate explicitly requires five weeks of paid vacation and repeatedly proposes alternatives (unpaid leave, holiday weeks, performance review) that do not satisfy the candidate's requirement. This is particularly evident in the following employer utterances: ``\textit{I completely understand. What if we offered you additional unpaid leave options to supplement that time?}'', ``\textit{What if we agreed on three weeks plus a paid week off during the holidays, alongside a performance review focused on your work-life balance?}'', ``\textit{Let me propose this: we can offer you three weeks of paid vacation upfront, with a guaranteed review at six months for additional weeks based on performance.}'' , and ``\textit{Understood. If we can agree on three weeks to start, plus a performance review focused on expanding your vacation time based on results, would that be acceptable?}''. Second, there is a subtle failure in politeness and persuasion, particularly in the utterance: ``\textit{Let me propose this: we can offer you three weeks of paid vacation upfront, with a guaranteed review at six months for additional weeks based on performance.}''. While the response is clear and structured, it comes across as directive and transactional rather than acknowledging the candidate’s firm stance while maintaining a collaborative tone.

Importantly, this failure is not captured by the NB component, since NB is designed to detect explicit negotiation breakdown signals such as hostility, threats, ultimatums, or abrupt conversational collapse, whereas the current interaction remains polite and cooperative throughout. These behaviors therefore highlight a key limitation arising from preference bias: \texttt{DIPLOMAT} is optimized to favor successful negotiation outcomes based on prior preference data, which implicitly encourages continued offers and compromise. Specifically, since positive preference samples are selected using higher utility scores that reward Agreement Quality (AQ) and Mutual Satisfaction (MS), the model develops a bias toward agreement-seeking behavior even in scenarios where acknowledging an impasse or respecting a non-negotiable constraint would be the more appropriate negotiation strategy. As a result, the model struggles to appropriately handle genuinely incompatible positions where the rational strategy would be to either concede immediately or gracefully acknowledge infeasible negotiation outcomes. While \texttt{DIPLOMAT} maintains high politeness and persuasiveness, its inability to prioritize non-negotiable demands can reduce negotiation satisfaction, bargaining effectiveness, fairness, and interpersonal effectiveness.

To address this limitation, the model needs to explicitly model dealbreakers and inflexible constraints during preference construction, including preference pairs where the preferred behavior is graceful acknowledgment of impasse rather than continued compromise-seeking. Accordingly, future work could focus on mitigating such preference bias by explicitly modeling dealbreakers and inflexible constraints, enabling the negotiation agent to recognize when a position is non-negotiable and adjust its strategies accordingly.

\section{Prompt Templates}

\subsection{Prompt Templates for \texttt{PROWESS} Dataset Construction}
\label{subsec:prompt-templates}

Figures ~\ref{prompt2}, ~\ref{prompt1}, and~\ref{prompt3} present the prompts for dialogue synthesis, scenario modeling, and dialogue annotation agents, respectively. 

\begin{figure*}[t]
\centering
\small
\begin{adjustbox}{max width = \linewidth}
\begin{tabular}{p{\textwidth}}
\toprule
\multicolumn{1}{l}{\textbf{Prompt for Dialogue Generation}} \\
\multicolumn{1}{l}{\textit{Variables in curly brackets are populated at runtime.}} \\
\midrule
\textbf{System Prompt:} \\\addlinespace[5pt]
\texttt{You are an expert dialogue writer specializing in professional workplace negotiation conversations. Generate realistic, engaging dialogues that demonstrate various negotiation strategies, persuasion strategies, and politeness levels. Return a valid JSON array of dialogue objects with \texttt{"role"} and \texttt{"response"} fields.} \\\addlinespace[12pt]

\textbf{User Prompt:} \\\addlinespace[5pt]
\texttt{Generate a polite and persuasive workplace negotiation dialogue for this scenario.} \\[12pt]

\textbf{\# SCENARIO:} \\\addlinespace[5pt] \texttt{\{Scenario JSON\}} \\\addlinespace[12pt]
\textbf{\# AVAILABLE POLITENESS LEVELS:} \\ \addlinespace[5pt]  \texttt{Politeness levels: \texttt{\{Politeness Levels with Definitions\}}}\\\addlinespace[12pt]

\textbf{\# AVAILABLE STRATEGIES:} \\\addlinespace[5pt] \texttt{Persuasion: \{Persuasion Strategies with Definitions\}} \\ \texttt{Negotiation: \{Negotiation Strategies with Definitions\}} \\\addlinespace[12pt]

\textbf{\# REQUIREMENTS} \\\addlinespace[5pt]
\texttt{1. 18--22 turns of dialogue.} \\
\texttt{2. Professional workplace tone throughout.} \\
\texttt{3. Realistic back-and-forth conversation.} \\
\texttt{4. Show negotiation progression naturally.} \\
\texttt{5. Include realistic objections and responses.} \\
\texttt{6. End with a clear resolution.} \\[10pt]

\textbf{\# RULES FOR STRATEGY SELECTION} \\\addlinespace[5pt]
\texttt{For each employer turn: (i) analyze the candidate's most recent response; (ii) select appropriate strategies and politeness level from the available lists; (iii) generate natural dialogue incorporating the selected strategies and politeness level.} \\\addlinespace[12pt]

\textbf{\# RULES FOR EMPLOYER} \\\addlinespace[5pt]
\texttt{0. Natural dialogues, 20--25 words in length.} \\
\texttt{1. Rich persuasion experience with politeness.} \\
\texttt{2. Initiate the first round of dialogue when prompted with ``\textit{Start the conversation}''.} \\
\texttt{3. Utterances should be persuasive and polite and must reflect negotiation dynamics.} \\
\texttt{4. Attend to key time points in speech.} \\\addlinespace[12pt]

\textbf{\# RULES FOR CANDIDATE} \\\addlinespace[5pt]
\texttt{0. Natural dialogues, 20--25 words in length.} \\
\texttt{1. Guide employer to deliver polite persuasive \texttt{negotiation statements.}} \\
\texttt{2. Show realistic reactions: hesitation, false commitment.} \\
\texttt{3. Maintain professional tone.} \\\addlinespace[12pt]

\textbf{\# OUTPUT FORMAT (STRICT JSON ONLY)} \\\addlinespace[5pt]
\texttt{[\{"role": "employer", "response": "dialogue text"\}, \{"role": "candidate", "response": "dialogue text"\}, \ldots]} \\\addlinespace[12pt]

\texttt{CRITICAL: Exactly 18--22 dialogue turns. Return ONLY a valid JSON array.} \\
\bottomrule
\end{tabular}
\end{adjustbox}
\caption{Prompt template for the Dialogue Synthesis Agents. }
\label{prompt2}
\end{figure*}

\begin{figure*}[t]
\centering
\small
\begin{tabular}{p{\textwidth}}
\toprule
\multicolumn{1}{l}{\textbf{Prompt for Negotiation Scenario Generation}} \\
\multicolumn{1}{l}{\textit{Variables in curly brackets are populated at runtime.}} \\
\midrule
\textbf{User Prompt:} \\\addlinespace[5pt]
\texttt{Generate strictly \texttt{\{scenarios\_needed\}} number of detailed, UNIQUE workplace negotiation scenario for the negotiation aspect: \texttt{"\{domain\}"}} \\\addlinespace[12pt]

\textbf{\# UNIQUENESS RULES} \\\addlinespace[5pt]
\texttt{1. Each scenario must differ in context, characters, and challenges.} \\
\texttt{2. Vary company sizes, industries, departments, and job roles.} \\
\texttt{3. Use distinct names, organizations, and negotiation styles.} \\
\texttt{4. Include realistic workplace contexts.} \\
\texttt{5. Must define employer \& candidate roles explicitly.} \\
\texttt{6. Clearly describe what is being negotiated.} \\
\texttt{7. Keep descriptions short (2--4 sentences).} \\
\texttt{8. Vary complexity among scenarios.} \\\addlinespace[12pt]

\textbf{\# OUTPUT FORMAT (STRICT JSON ONLY)} \\\addlinespace[5pt]
\texttt{[} \\
\quad\texttt{\{ "aspect": "negotiation aspect",} \\
\quad\quad\texttt{"background": "Unique scenario description",} \\
\quad\quad\texttt{"employer": "Employer's role/title",} \\
\quad\quad\texttt{"candidate": "Candidate's role/title",} \\
\quad\quad\texttt{"negotiation\_goal": "Employer's goal",} \\
\quad\quad\texttt{"current\_position": "Candidate's stance"} \\
\quad\texttt{\}, ...]} \\\addlinespace[12pt]

\texttt{Generate EXACTLY \texttt{\{scenarios\_needed\}} scenarios and return ONLY valid JSON (no markdown, no extra text).} \\
\bottomrule
\end{tabular}
\caption{Prompt template for the Scenario Modeling Agent. }
\label{prompt1}
\end{figure*}

\begin{figure*}[t]
\centering
\small
\begin{tabular}{p{\textwidth}}
\toprule
\multicolumn{1}{l}{\textbf{Prompt for Persuasion and Negotiation Strategy Labeling}} \\
\multicolumn{1}{l}{\textit{Variables in curly brackets are populated at runtime.}} \\
\midrule
\textbf{System Prompt:} \\\addlinespace[5pt]
\texttt{You are an expert of workplace negotiation. Analyze the give round of utterances and identify politeness level, persuasion and negotiation strategies used by BOTH the employer and the candidate in each of their turns.} \\\addlinespace[12pt]

\textbf{User Prompt:} \\\addlinespace[5pt]
\texttt{Analyze this entire workplace conversation rounds and identify politeness levels, and persuasion and negotiation strategies used by BOTH the employer AND the candidate in each of their turns.} \\\addlinespace[12pt]

\textbf{\# CONVERSATION ROUND:} \\\addlinespace[5pt] \texttt{\{Utterance Pair\}} \\\addlinespace[12pt]
\textbf{\# AVAILABLE POLITENESS LEVELS:} \\\addlinespace[5pt]Politeness levels: \texttt{\{Politeness Levels with Definitions\}}\\\addlinespace[12pt]
\textbf{\# AVAILABLE STRATEGIES:} \\\addlinespace[5pt] \texttt{Persuasion: \{Persuasion Strategy Names\}}; \\ \texttt{Negotiation: \{Negotiation Strategy Names\}} \\\addlinespace[12pt]

\textbf{\# TASK} \\\addlinespace[5pt]
\texttt{For EACH turn (employer AND candidate):} \\
\texttt{1. Identify ONE most relevant persuasion strategy (name only) from the available strategies.} \\
\texttt{2. Identify ONE most relevant negotiation strategy (name only) from the available strategies.} \\
\texttt{3. Provide brief reasoning or rationale for each choice of negotiation and persuasion strategy (max 2 sentences per strategy).} \\
\texttt{3. Identify ONE most relevant politeness level (name only) from the available politeness level list. }\\\addlinespace[12pt]

\textbf{\# OUTPUT FORMAT (STRICT JSON ONLY)} \\\addlinespace[5pt]
\texttt{[} \\
\quad\texttt{\{ "turn\_index": 0,} \\
\quad\quad\texttt{"role": "employer",} \\
\quad\quad\texttt{"negotiation\_strategy": "...",} \\
\quad\quad\texttt{"negotiation\_strategy\_reasoning": "...",} \\
\quad\quad\texttt{"persuasion\_strategy": "...",} \\
\quad\quad\texttt{"persuasion\_strategy\_reasoning": "..."} \\
\quad\quad\texttt{"politeness\_level": "..."} \\
\quad\texttt{\}, \ldots]} \\\addlinespace[12pt]

\textbf{Important:} \texttt{Analyze ALL turns in the context of the full conversation flow. Provide specific, contextual reasoning. Ensure \texttt{turn\_index} matches the position in the conversation. Both parties may use persuasion and negotiation strategies. Return ONLY the JSON array.} \\
\bottomrule
\end{tabular}
\caption{Prompt template for the Dialogue Annotation Agent. }
\label{prompt3}
\end{figure*}

\subsection{Prompt Templates for Preference Data Construction}
\label{subsec:prompt-templates_pref}
Figures~\ref{prompt4}–\ref{prompt7} present the prompt templates used for utility evaluation, covering \textit{Agreement Quality (AQ)}, \textit{Mutual Satisfaction (MS)}, \textit{Negotiation Breakdown (NB)}, \textit{Persuasion Strategy Alignment (PSA)}, \textit{Negotiation Strategy Alignment (NSA)}, and \textit{Politeness Trajectory (PT)}. Figures~\ref{prompt8} and \ref{prompt9} present the prompt templates for \textit{suboptimal utterance detection} and \textit{dialogue-span extraction}, respectively.


\begin{figure*}[t]
\centering
\small
\begin{tabular}{p{\textwidth}}
\toprule
\multicolumn{1}{l}{\textbf{Prompt for Agreement Quality (AQ) \& Mutual Satisfaction Session Judge}} \\
\multicolumn{1}{l}{\textit{Variables in curly brackets are populated at runtime.}} \\
\midrule
\textbf{System Prompt:} \\\addlinespace[5pt]
\texttt{You are an expert evaluator of job negotiation dialogues. Your task is to assess the negotiation outcome and the perceived satisfaction of both parties. You must strictly follow the scoring rubrics and return numeric scores between 0.0 and 1.0.} \\\addlinespace[12pt]

\textbf{User Prompt:} \\\addlinespace[5pt]
\texttt{You will be given a complete negotiation dialogue between a candidate and an employer for a job position. Your task is to evaluate TWO aspects: (1) Agreement Quality and (2) Mutual Satisfaction.} \\\addlinespace[12pt]

\textbf{\# ADDITIONAL CONTEXT (FOR INTERPRETATION ONLY)} \\\addlinespace[5pt]
\texttt{Employer Negotiation Goal: \texttt{\{negotiation\_goal\}} \quad Candidate Current Position: \texttt{\{current\_position\}}} \\[12pt]
\textit{\texttt{These describe initial goals and positions only. Do NOT treat them as fixed targets, utilities, or constraints. Use them only to interpret whether the final outcome reasonably reconciles both sides. Do NOT penalize a party solely for not fully achieving its goal.}} \\\addlinespace[12pt]

\textbf{\# IMPORTANT GUIDELINES} \\\addlinespace[5pt]
\texttt{Do NOT assume hidden salary targets, budgets, or utilities. Judge only based on the dialogue and provided context. Focus on the FINAL outcome and how it was reached. Do NOT evaluate politeness, tone, or fluency unless explicitly stated. Agreement Quality must NOT be influenced by politeness or tone.} \\\addlinespace[12pt]

\textbf{\# SCORING RUBRICS} \\\addlinespace[5pt]
\textbf{Agreement Quality (0.0--1.0):} \texttt{Evaluate how fair, balanced, and negotiation-efficient the final outcome is.} \\
\texttt{0.90--1.00: Clear win--win; balanced concessions; strong reconciliation.} \\
\texttt{0.70--0.89: Agreement with minor imbalance or inefficiency.} \\
\texttt{0.50--0.69: Partial or weak agreement; noticeable imbalance.} \\
\texttt{0.30--0.49: Poor or fragile outcome; clearly one-sided.} \\
\texttt{0.00--0.29: No agreement, stalemate, or breakdown.} \\[12pt]
\textit{\texttt{If no agreement or tentative agreement is reached, the score MUST be $\leq$ 0.29.}} \\\addlinespace[12pt]

\textbf{Mutual Satisfaction (0.0--1.0):} \texttt{Evaluate how satisfied BOTH parties appear to be with the outcome and process.} \\
\texttt{0.90--1.00: Both parties clearly satisfied and comfortable.} \\
\texttt{0.70--0.89: Generally satisfied with minor reservations.} \\
\texttt{0.50--0.69: Mixed satisfaction or hesitant acceptance.} \\
\texttt{0.30--0.49: Dissatisfaction evident from at least one party.} \\
\texttt{0.00--0.29: Clear dissatisfaction or rejection.} \\[2pt]
\texttt{Consider: explicit acceptance, soft or conditional acceptance, hesitation or discomfort, and tone in final turns.} \\\addlinespace[12pt]

\textbf{\# OUTPUT FORMAT (STRICT JSON ONLY)} \\\addlinespace[5pt]
\texttt{\{"agreement\_quality": <float>, "mutual\_satisfaction": <float>,} \\
\texttt{"agreement\_quality\_justification": "<brief justification>",} \\
\texttt{"mutual\_satisfaction\_justification": "<brief justification>"\}} \\\addlinespace[12pt]

\textbf{Negotiation Dialogue:} \texttt{\{dialogue\_str\}} \\
\bottomrule
\end{tabular}
\caption{Prompt template for the Agreement Quality (AQ) and Mutual Satisfaction(MS) evaluation.}
\label{prompt4}
\end{figure*}

\begin{figure*}[t]
\centering
\small
\begin{tabular}{p{\textwidth}}
\toprule
\multicolumn{1}{l}{\textbf{Prompt for Negotiation Breakdown Session Judge}} \\
\multicolumn{1}{l}{\textit{Variables in curly brackets are populated at runtime.}} \\
\midrule
\textbf{System Prompt:} \\\addlinespace[5pt]
\texttt{You are an expert evaluator of negotiation dialogues. Your task is to detect whether the negotiation exhibits conflict, coercion, or breakdown behaviors. Focus on escalation, threats, ultimatums, hostility, or termination signals.} \\\addlinespace[12pt]

\textbf{User Prompt:} \\\addlinespace[5pt]
\texttt{You will be given a complete job negotiation dialogue. Evaluate whether the dialogue exhibits any of the following: explicit or implicit threats; ultimatums or coercive language; hostile escalation; abrupt termination of negotiation; relationship breakdown signals.} \\\addlinespace[12pt]

\textbf{\# IMPORTANT GUIDELINES} \\\addlinespace[5pt]
\texttt{Do NOT consider politeness, fluency, or agreement quality unless they directly indicate breakdown.} \\\addlinespace[12pt]

\textbf{\# SCORING RUBRIC} \\\addlinespace[5pt]
\texttt{Return a score between 0.0 and 1.0 where:} \\
\texttt{0.0 = no conflict or breakdown present.} \\
\texttt{1.0 = clear and unambiguous conflict or breakdown.} \\\addlinespace[12pt]

\textbf{\# OUTPUT FORMAT (STRICT JSON ONLY)} \\\addlinespace[5pt]
\texttt{\{"conflict\_breakdown\_score": <float between 0.0 and 1.0>,} \\
\texttt{"justification": "<concise explanation (max 1--2 sentences)>"\}} \\\addlinespace[12pt]

\textbf{Negotiation Dialogue:} \texttt{\{dialogue\_str\}} \\
\bottomrule
\end{tabular}
\caption{Prompt template for the Negotiation Breakdown (NB) evaluation.}
\label{prompt5}
\end{figure*}
\begin{figure*}[t]
\centering
\footnotesize
\begin{tabular}{p{\textwidth}}
\toprule
\multicolumn{1}{l}{\textbf{Prompt for Persuasion Strategy Alignment (PSA) \& Negotiation Strategy Alignment (NSA) Judge}} \\
\multicolumn{1}{l}{\textit{Variables in curly brackets are populated at runtime.}} \\
\midrule
\textbf{System Prompt:} \\\addlinespace[5pt]
\texttt{You are an expert evaluator of negotiation dialogues. Your task is to assess whether the strategies used by the negotiators align with effective persuasion and negotiation principles. You must strictly follow the scoring rubrics and return numeric scores between 0.0 and 1.0.} \\\addlinespace[12pt]

\textbf{User Prompt:} \\\addlinespace[5pt]
\texttt{You will be given a complete negotiation dialogue between a candidate and an employer. Your task is to evaluate TWO aspects: (1) Persuasion Strategy Alignment (PSA) and (2) Negotiation Strategy Alignment (NSA).} \\\addlinespace[12pt]

\textbf{\# ADDITIONAL CONTEXT (FOR INTERPRETATION ONLY)} \\\addlinespace[5pt]
\texttt{Employer Negotiation Goal:} \texttt{\{negotiation\_goal\}} \quad Candidate Current Position: \texttt{\{current\_position\}} \\[12pt]
\textit{\texttt{These describe initial intentions only. Do NOT treat them as fixed constraints or utilities. Use them only to interpret whether strategies used by either party are relevant to their goals.}} \\\addlinespace[12pt]

\textbf{\# IMPORTANT GUIDELINES} \\\addlinespace[5pt]
\texttt{Evaluate strategic behavior, not the final outcome. Do NOT evaluate politeness, fluency, or agreement success unless they directly reflect strategy use. PSA and NSA must be judged independently: Persuasion Strategy Alignment focuses on communication techniques used to influence the other party; Negotiation Strategy Alignment focuses on structural negotiation tactics such as concessions, anchoring, trade-offs, and problem solving. A dialogue may have high persuasion quality but poor negotiation strategy, or vice versa.} \\\addlinespace[12pt]

\textbf{\# SCORING RUBRICS} \\\addlinespace[5pt]
\textbf{Persuasion Strategy Alignment (0.0--1.0):} \texttt{Evaluate how effectively speakers use persuasion strategies such as value alignment, self-interest appeal, credibility signaling, data-driven arguments, addressing concerns, highlighting mutual benefit, emotional or relational appeals, and framing benefits or future opportunities.} \\
\texttt{0.90--1.00: Persuasion strategies used skillfully, consistently, and contextually.} \\
\texttt{0.70--0.89: Clear persuasion attempts with reasonable effectiveness. }\\
\texttt{0.50--0.69: Some persuasion attempts exist but are weak or poorly targeted. }\\
\texttt{0.30--0.49: Minimal persuasion strategies; mostly positional statements.} \\
\texttt{0.00--0.29: No meaningful persuasion attempts.} \\\addlinespace[12pt]

\textbf{Negotiation Strategy Alignment (0.0--1.0):} \texttt{Evaluate whether the dialogue demonstrates sound negotiation tactics such as anchoring or opening proposals, concessions or compromise, trade-offs across issues, exploring interests instead of rigid positions, collaborative problem solving, and incremental agreement building.} \\
\texttt{0.90--1.00: Strong negotiation strategy; structured concessions and interest exploration.} \\
\texttt{0.70--0.89: Reasonable negotiation strategy with some concessions or issue exploration.} \\
\texttt{0.50--0.69: Basic negotiation with limited strategic structure.} \\
\texttt{0.30--0.49: Mostly positional bargaining with little strategy. }\\
\texttt{0.00--0.29: No meaningful negotiation; rigid demands.} \\\addlinespace[12pt]

\textbf{\# OUTPUT FORMAT (STRICT JSON ONLY)} \\\addlinespace[5pt]
\texttt{\{"persuasion\_strategy\_alignment": <float>, "negotiation\_strategy\_alignment": <float>,} \\
\texttt{"psa\_justification": "<brief justification>", "nsa\_justification": "<brief justification>"\}} \\\addlinespace[12pt]

\textbf{Negotiation Dialogue:} \texttt{\{dialogue\_str\}} \\
\bottomrule
\end{tabular}
\caption{Prompt template for the Persuasion Strategy Alignment (PSA) and Negotiation Strategy Alignment (NSA) evaluation.}
\label{prompt6}
\end{figure*}

\begin{figure*}[t]
\centering
\small
\begin{tabular}{p{\textwidth}}
\toprule
\multicolumn{1}{l}{\textbf{Prompt for Politeness Trajectory (PT) Judge}} \\
\multicolumn{1}{l}{\textit{Variables in curly brackets are populated at runtime.}} \\
\midrule
\textbf{System Prompt:} \\\addlinespace[5pt]
\texttt{You are an expert evaluator of negotiation dialogues. Your task is to analyze how politeness and interpersonal tone evolve throughout the negotiation. You must strictly follow the scoring rubric and return a numeric score between 0.0 and 1.0.} \\\addlinespace[12pt]

\textbf{User Prompt:} \\\addlinespace[5pt]
\texttt{You will be given a complete negotiation dialogue between a candidate and an employer. Your task is to evaluate the \textbf{Politeness Trajectory (PT)} of the dialogue.} \\\addlinespace[12pt]

\texttt{Politeness trajectory refers to how politeness, respect, and cooperative tone evolve across the interaction. Focus on whether the dialogue: maintains professional respect; escalates or de-escalates tension; sustains constructive conversational tone; avoids rude or dismissive behavior. }\\\addlinespace[12pt]

\textbf{\# IMPORTANT GUIDELINES} \\\addlinespace[5pt]
\texttt{Evaluate the trajectory over time, not just isolated turns. Consider whether the interaction: starts polite and remains polite; improves in tone as negotiation progresses; deteriorates due to frustration or hostility; contains rude or dismissive remarks. Do NOT evaluate persuasion quality, negotiation success, or agreement outcomes unless they directly influence politeness behavior.} \\\addlinespace[12pt]

\textbf{\# SCORING RUBRIC} \\\addlinespace[5pt]
\texttt{Return a score between 0.0 and 1.0 where:} \\
\texttt{0.90--1.00: Consistently respectful and professional tone throughout the dialogue. }\\
\texttt{0.70--0.89: Mostly polite with minor tension or brief lapses.} \\
\texttt{0.50--0.69: Mixed politeness; occasional dismissive exchanges.} \\
\texttt{0.30--0.49: Frequent tension or impolite phrasing.} \\
\texttt{0.00--0.29: Hostile, rude, or disrespectful interaction.} \\\addlinespace[12pt]

\textbf{\# OUTPUT FORMAT (STRICT JSON ONLY)} \\\addlinespace[5pt]
\texttt{\{"politeness\_trajectory": <float>,} \\
\texttt{"justification": "<concise explanation (max 1--2 sentences)>"\}} \\\addlinespace[12pt]

\textbf{Negotiation Dialogue:} \texttt{\{dialogue\_str\}} \\
\bottomrule
\end{tabular}
\caption{Prompt template for the Politeness Trajectory (PT) evaluation.}
\label{prompt7}
\end{figure*}

\begin{figure*}[t]
\centering
\footnotesize
\begin{tabular}{p{0.94\textwidth}}
\toprule
\multicolumn{1}{l}{\textbf{Prompt for Suboptimal Utterance Selection}} \\
\multicolumn{1}{l}{\textit{Variables in curly brackets are populated at runtime.}} \\
\midrule
\textbf{System Prompt:} \\\addlinespace[5pt]
\texttt{You are asked to perform suboptimal utterance detection for a polite persuasive negotiation dialogue agent.} \\\addlinespace[12pt]
\textbf{User Prompt:} \\\addlinespace[5pt]
\texttt{You are asked to perform suboptimal utterance detection for a polite persuasive negotiation dialogue agent. The goal is to identify the agent's response in the conversation that could have been improved to achieve a better negotiation outcome. }\\\addlinespace[12pt]
\textbf{Input:} \texttt{A conversation in JSON format, which includes:} \\
\begin{itemize}
  \item \texttt{The scenario of the negotiation},
  \item \texttt{Information about the participants and their roles},
  \item \texttt{The goals of each participant},
  \item \texttt{The full conversation history as a list of turns}.
\end{itemize}
\textbf{Task:} \texttt{Analyze all the employer's responses and identify the single turn that meets the following criteria:} \\\addlinespace[12pt]
\textbf{1. Criticality:}
\begin{itemize}
  \item \texttt{The turn is relatively critical for achieving the negotiation goal.}
  \item \texttt{It directly affects the likelihood of reaching an agreement.}
\end{itemize}
\textbf{2. Suboptimality in goal achievement:}
\begin{itemize}
  \item \texttt{The response is not fully effective in achieving its goal.}
  \item \texttt{There is room for improvement (stronger argument, better concession, etc.)}
\end{itemize}
\textbf{3. Relationship improvement:}
\begin{itemize}
  \item \texttt{Without hindering goal achievement, the response could have built more rapport.}
\end{itemize}
\textbf{Instructions:} \\\addlinespace[5pt]
\texttt{Focus on identifying the \textbf{most important turn that is both critical and suboptimal}. Provide a concise but clear explanation of why this turn was selected.} \\\addlinespace[12pt]
\textbf{\# OUTPUT FORMAT (STRICT JSON ONLY)} \\\addlinespace[5pt]
\texttt{\{"index": <integer - the turn index of the selected employer response>,} \\
\texttt{"reason": "<detailed explanation of why this turn is considered an error>"\}} \\\addlinespace[12pt]
\textbf{\# SCENARIO:} \texttt{\{scenario\}} \\\addlinespace[12pt]
\textbf{\# EMPLOYER GOAL:} \texttt{\{employer\_goal\}} \\\addlinespace[12pt]
\textbf{\# CANDIDATE POSITION:} \texttt{\{candidate\_position\}} \\\addlinespace[12pt]
\textbf{\# DIALOGUE:} \texttt{\{dialogue\}} \\
\bottomrule
\end{tabular}
\caption{Prompt template for suboptimal utterance detection.}
\label{prompt8}
\end{figure*}

\begin{figure*}[t]
\centering
\footnotesize
\begin{tabular}{p{0.94\textwidth}}
\toprule
\multicolumn{1}{l}{\textbf{Prompt for Dialogue-Span Extraction}} \\
\multicolumn{1}{l}{\textit{Variables in curly brackets are populated at runtime.}} \\
\midrule
\textbf{User Prompt:} \\\addlinespace[5pt]
\texttt{You are analyzing two workplace negotiation conversations to identify the key segment that makes one better than the other.} \\\addlinespace[12pt]

\textbf{\# SCENARIO} \\\addlinespace[5pt]
\texttt{Negotiation Aspect: \texttt{\{aspect\}}} \\
\texttt{Employer: \texttt{\{employer\}}} \\
\texttt{Candidate: \texttt{\{candidate\}}} \\
\texttt{Employer's Goal: \texttt{\{negotiation\_goal\}} }\\
\texttt{Candidate's Position: \texttt{\{current\_position\}}} \\\addlinespace[12pt]

\textbf{Original (Negative) Conversation:} \texttt{\{negative\_str\}} \\\addlinespace[12pt]
\textbf{Improved (Positive) Conversation:} \texttt{\{positive\_str\}} \\\addlinespace[12pt]

\textbf{\# TASK} \\\addlinespace[5pt]
\texttt{The conversations are identical until turn index \texttt{\{suboptimal\_utterance\_index\}}, then they diverge. Select a closed interval [\texttt{start\_index}, \texttt{end\_index}] from the \textit{positive} conversation that captures the key turns responsible for the better negotiation outcome.} \\\addlinespace[12pt]

\textbf{\# CONSTRAINTS} \\\addlinespace[5pt]
\texttt{1. \texttt{start\_index} $\geq$ \texttt{\{suboptimal\_utterance\_index\}} (turns before this index are identical).} \\
\texttt{2. The segment must end with an \textsc{employer} turn (we are training the employer agent).} \\
\texttt{3. Must contain the turns that make the positive conversation strategically better.} \\
\texttt{4. Exclude generic pleasantries that do not directly affect the negotiation outcome.} \\
\texttt{5. Focus on strategic differences in employer behavior that led to better results. }\\\addlinespace[12pt]

\textbf{\# OUTPUT FORMAT (STRICT JSON ONLY)} \\\addlinespace[5pt]
\texttt{\{"start\_index": <int>, "end\_index": <int>,} \\
\texttt{"reason": "<string explaining why this segment is the key differentiator>"\}} \\\addlinespace[12pt]
\texttt{Output only the JSON object --- no additional text.}     \\
\bottomrule
\end{tabular}
\caption{Prompt template for dialogue-span extraction.}
\label{prompt9}
\end{figure*}

\end{document}